\documentclass[lettersize,journal]{IEEEtran}
\usepackage{amsmath,amsfonts}
\usepackage{algorithmic}
\usepackage{algorithm}
\usepackage{array}
\usepackage[caption=false,font=normalsize,labelfont=sf,textfont=sf]{subfig}
\usepackage{textcomp}
\usepackage{stfloats}
\usepackage{url}
\usepackage{verbatim}
\usepackage{graphicx}
\usepackage{cite}
\usepackage{capt-of}
\usepackage{hyperref}
\hypersetup{
    colorlinks=true,
    linkcolor=red,
    filecolor=magenta,      
    urlcolor=cyan,
    pdftitle={Overleaf Example},
    pdfpagemode=FullScreen,
    }
\usepackage[dvipsnames]{xcolor}
\usepackage{colortbl}
\usepackage{multirow}
\usepackage{adjustbox}
\usepackage{amsmath}
\newcommand{\ie}{\textit{i.e.}}
\newcommand{\eg}{\textit{e.g.}}
\newcommand{\vs}{\textit{vs.}}
\usepackage{pifont}
\newcommand{\cmark}{\ding{51}}
\newcommand{\xmark}{\ding{55}}
\definecolor{third}{rgb}{1,1, 0.6}
\definecolor{second}{rgb}{1, 0.8, 0.6}
\definecolor{best}{rgb}{1, 0.6, 0.6}
\newcommand{\cb}{\cellcolor{best}}
\newcommand{\cs}{\cellcolor{second}}

\usepackage[switch]{lineno}

\begin{document}
    


\title{GD$^2$-NeRF: Generative Detail Compensation via GAN and Diffusion for One-shot Generalizable Neural Radiance Fields}

\author{
        Xiao Pan, Zongxin Yang$^*$, Shuai Bai,  Yi Yang
            }
            

\twocolumn[{%
\renewcommand\twocolumn[1][]{#1}%
    \maketitle        
    \centering
    \vspace*{-8mm}
    \includegraphics[width=0.9\textwidth]{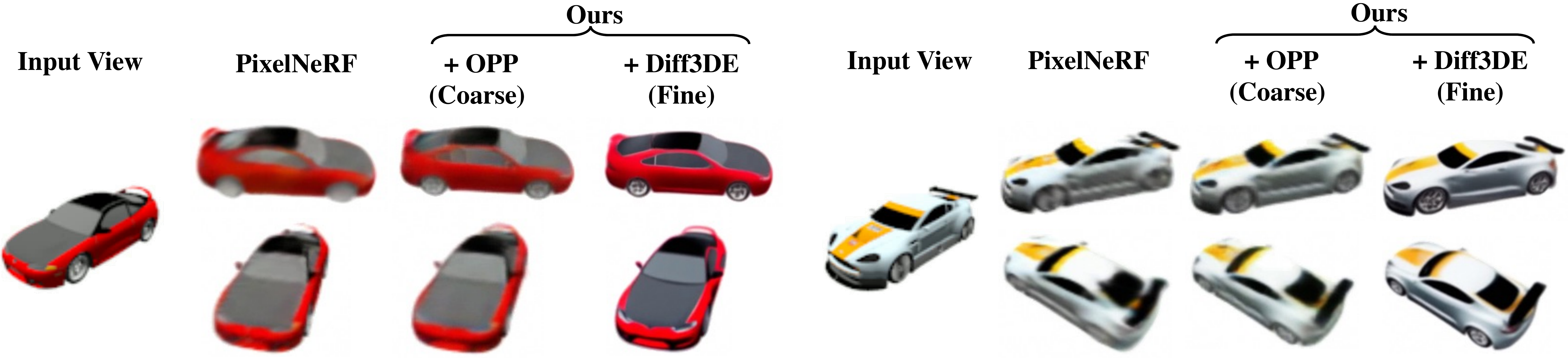}
    \vspace*{-0.2cm}
    \captionof{figure}{Given a single reference image, our method GD$^2$-NeRF  synthesizes novel views with \textbf{vivid plausible details} in an \textbf{inference-time finetuning-free} manner. It is a coarse-to-fine generative detail compensation framework composed of OPP and Diff3DE. OPP first injects the GAN model into existing OG-NeRF pipelines, \eg, PixelNeRF~\cite{PixelNeRF_yu2021pixelnerf}, for achieving in-distribution detail priors. Then, Diff3DE further incorporates the out-distribution detail priors from the pre-trained diffusion models~\cite{rombach2022LatentDiff,zhang2023addingControlNet}. \textcolor{red}{\textbf{We highly recommend readers to check our video demos for more intuitive comparisons.}}}\label{fig:teaser}
    }]

\markboth{Journal of \LaTeX\ Class Files,~Vol.~14, No.~8, August~2021}%
{Shell \MakeLowercase{\textit{et al.}}: A Sample Article Using IEEEtran.cls for IEEE Journals}


\let\thefootnote\relax\footnotetext{Xiao Pan, Zongxin Yang, and Yi Yang are with the ReLER Lab, CCAI, Zhejiang University, Hangzhou, 310000, China, and Shuai Bai is with the Alibaba DAMO Academy, Hangzhou, 310000, China. (email: xiaopan@zju.edu.cn, yangzongxin@zju.edu.cn, baishuai.bs@alibaba-inc.com, 
yangyics@zju.edu.cn) \par $*$ Corresponding author. }

\begin{abstract}
In this paper, we focus on the One-shot Novel View Synthesis (O-NVS) task which targets synthesizing photo-realistic novel views given only one reference image per scene. Previous One-shot Generalizable Neural Radiance Fields (OG-NeRF) methods solve this task in an inference-time finetuning-free manner, yet suffer the blurry issue due to the encoder-only architecture that highly relies on the limited reference image. On the other hand, recent diffusion-based image-to-3d methods show vivid plausible results via distilling pre-trained 2D diffusion models into a 3D representation, yet require tedious per-scene optimization. Targeting these issues, we propose the GD$^2$-NeRF, a Generative Detail compensation framework via GAN and Diffusion that is both inference-time finetuning-free and with vivid plausible details. 
In detail, following a coarse-to-fine strategy, GD$^2$-NeRF is mainly composed of a One-stage Parallel Pipeline (OPP) and a 3D-consistent Detail Enhancer (Diff3DE). At the coarse stage, OPP first efficiently inserts the GAN model into the existing OG-NeRF pipeline for primarily relieving the blurry issue with in-distribution priors captured from the training dataset, achieving a good balance between sharpness  (LPIPS, FID) and fidelity (PSNR, SSIM). Then, at the fine stage, Diff3DE further leverages the pre-trained image diffusion models to complement rich out-distribution details while maintaining decent 3D consistency.
Extensive experiments on both the synthetic and real-world datasets show that GD$^2$-NeRF noticeably improves the details while without per-scene finetuning.
\vspace{-2mm}

\end{abstract}

\begin{IEEEkeywords}
One-shot novel view synthesis, generalizable neural radiance fields, 
3D reconstruction, GAN, diffusion model.
\end{IEEEkeywords}

\section{\textbf{Introduction}}
\label{sec:intro}

One-shot Novel View Synthesis (O-NVS) is a long-standing problem in computer vision and graphics which targets on synthesizing photo-realistic novel views of a scene given a single reference image. An important technology solving this task is the One-shot Generalizable Neural Radiance Fields (OG-NeRF) which trains image-conditioned NeRF across scenes for learning general 3D priors and can generalize to a new scene by a single feed-forward pass, \ie, \textit{inference-time finetuning-free}.

\begin{figure}[t]
\setlength{\abovecaptionskip}{0cm}
\small
\centering
\includegraphics[width=\linewidth]{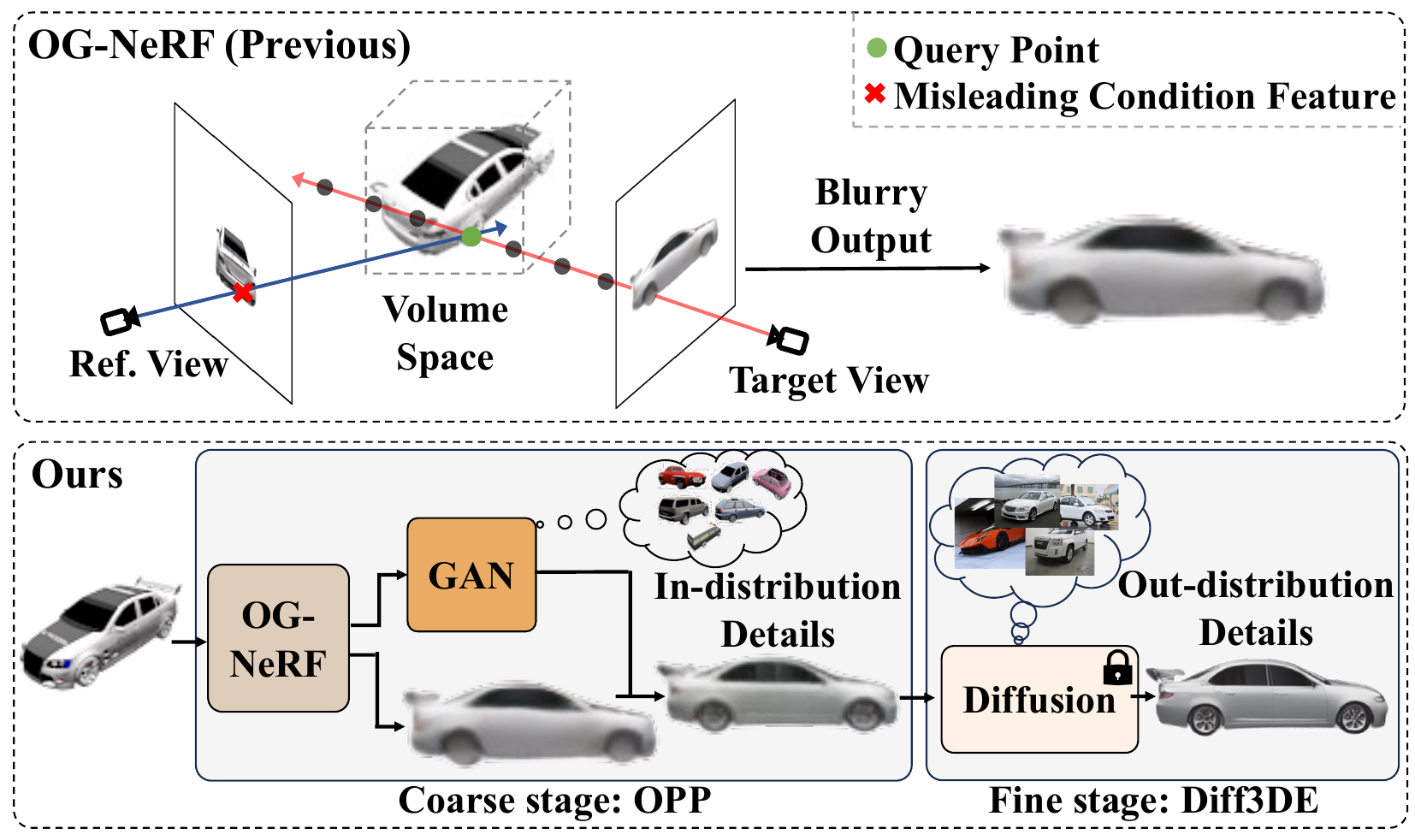}
\caption{\textbf{Comparison between the existing encoder-only OG-NeRF and our generative detail compensation perspective (\S~\ref{sec:intro}).} OG-NeRF suffers the blurry issue due to the projected misleading features while we propose to complement the object details via the prior learned by the generative model.}
\label{fig_idea}
\vspace{-6mm}
\end{figure}

However, 1) the existing OG-NeRF methods~\cite{PixelNeRF_yu2021pixelnerf,lin2023visionnerf,FE_NVS_guo2022fast} mainly suffer the \textit{blurry issue} since their encoder-only architectures highly rely on the reference images that contain limited information. 
For instance, ~\cite{PixelNeRF_yu2021pixelnerf,lin2023visionnerf} first encodes the reference image into a 2D feature map and then indexes the condition features via pixel-wise projection. It works well when the target view is close to the reference view, yet tends to get blurry as the view difference becomes larger since the reference image can provide limited or even misleading scene information, \eg, as shown by the upper row of Fig.~\ref{fig_idea}, the query point requires the wheel information from the right view while the misleading body information from the back view is projected. 2) On the other hand, recent advances of diffusion-based image-to-3d methods~\cite{liu2023zero123,tang2023makeit3d} show vivid plausible novel view results via distilling the 2D generative priors from pre-trained diffusion models~\cite{rombach2022LatentDiff,zhang2023addingControlNet} into a 3D representation, yet requires tedious \textit{per-scene optimization}.

Targeting these issues, we explore an O-NVS framework that is both \textit{inference-time finetuning-free }and with \textit{vivid plausible outputs}. To this end, we propose the GD$^2$-NeRF, a coarse-to-fine generative detail compensation framework that hierarchically includes GAN and pre-trained diffusion models into OG-NeRF. GD$^2$-NeRF is mainly composed of a {One-stage Parallel Pipeline (OPP)} that captures in-distribution priors via GAN model and a {Diffusion-based 3D-consistent Enhancer (Diff3DE)} that injects out-distribution priors from pre-trained diffusion models~\cite{rombach2022LatentDiff,zhang2023addingControlNet}, as illustrated by the second row of Fig.~\ref{fig_idea}. In detail:

(i) \textit{Coarse-stage OPP (GAN model)}. At the coarse stage, we intend to devise a pipeline that efficiently injects the GAN model into the existing OG-NeRF pipeline to primarily relieve the blurry issue using in-distribution detail priors captured from the training dataset. 
Targeting this, a naive solution is to directly build the GAN model on top of the OG-NeRF model tandemly, either in a two-stage or one-stage manner, as illustrated by the first two rows in Fig. \ref{fig_simple_pipelines} and will be detailed in~\S~\ref{sec:coarse_stage_opp}. However, though the \textit{sharpness} (LPIPS, FID) is significantly improved,  we empirically find it hard to maintain the \textit{fidelity} (PSNR, SSIM), even with the more coherent one-stage tandem pipeline. 

To address such contradiction between \textit{fidelity} and \textit{sharpness}, 
we further propose the One-stage Parallel Pipeline (OPP) that integrates the OG-NeRF and GAN model in a unified parallel framework, as shown by the bottom part of Fig. \ref{fig_simple_pipelines}. It is built on the one-stage tandem pipeline with the proposed  Dual-Paradigm Structure (DPS), Confidence Radiance Fields (CoRF), and Dual-Paradigm Fusion (DPF).
With DPS, the OG-NeRF model and the GAN model can be optimized in parallel within a single framework. Then, CoRF takes the occlusion information as input and predicts a confidence map which adaptively gives the blurry part with low confidence. Finally, DPF aggregates the outputs from two paradigms via the learned confidence map.

(ii) \textit{Fine-stage Diff3DE (diffusion model).}  Due to the limited size and quality of the training datasets, we find the in-distribution prior at the coarse stage is not enough for vivid outputs with rich plausible details. Therefore, at the fine stage, to break through such limitation, we turn to the large-scale diffusion models~\cite{rombach2022LatentDiff,zhang2023addingControlNet} pre-trained on billions of high-quality images for more vivid out-distribution details. 

However, naively using such an image diffusion model to process the rendered views from the coarse stage frame-by-frame leads to poor 3D consistency. 
Targeting this issue and inspired by the recent advances in zero-shot diffusion-based video editing methods~\cite{yang2023rerender,geyer2023tokenflow}, we propose the Diffusion-based 3D Enhancer (Diff3DE). The main idea of Diff3DE is to first ensure the consistency between several keyframes dispersed around the dome. Then, given an arbitrary target view, the information from nearby keyframes is aggregated in the feature space based on view information, formulating a robust 3D-consistent enhancer.

Extensive experiments on both the synthetic and real-world datasets show that, without any inference-time finetuning, 1) our OPP shows noticeable improvements over the baseline methods with balanced sharpness and fidelity while with little additional cost, and 2) Diff3DE can further compensate rich plausible details with decent 3D-consistency. 

Our contributions are summarized as follows:
\begin{itemize}

    \item  We devise a coarse-to-fine generative detail compensation framework, GD$^2$-NeRF, for O-NVS task that is both inference-time finetuning-free and with vivid plausible outputs.

    \item Our coarse-stage method OPP (\S~\ref{sec:coarse_stage_opp}) effectively inserts the GAN model into the existing OG-NeRF pipeline to primarily relieve the blurry issue with a good balance between fidelity and sharpness.

    \item To our best knowledge, our fine-stage method Diff3DE (\S~\ref{fine_stage_Diff3DE}) makes the early attempt to directly use the pre-trained diffusion model as a 3D-consistent enhancer without any further finetuning.
 
\end{itemize}

\section{\textbf{Related Works}}
\label{sec:related_work}
\subsection{\textbf{One-shot Novel View Synthesis}}
Recently, with the rapid development of the 3D computer vision community, there exist different technologies that can solve the one-shot novel view synthesis (O-NVS) task, though under different settings, including {OG-NeRF}~\cite{PixelNeRF_yu2021pixelnerf,FE_NVS_guo2022fast,lin2023visionnerf,gu2023nerfdiff}, {Geometry-free Methods}~\cite{sitzmann2021LFN,srt22,watson20223DiM}, {3D GAN}~\cite{PiGAN_chan2021pi,Pix2NeRF_cai2022pix2nerf,chan2022EG3D},  and {Large-model-based Image-to-3D}~\cite{Xu_2022_SinNeRF,Jain_2021_ICCV_dietnerf,liu2023zero123,tang2023makeit3d}. 

Our work is motivated from the OG-NeRF perspective yet not limited by it. Specifically, we target relieving the blurry issue of existing OG-NeRF methods while maintaining its nice property of \textit{inference-time finetuning-free}. However, in contrast to previous OG-NeRF methods that mainly focus on improving the in-distribution details from the limited dataset, we make the early attempt to also include the out-distribution details from the powerful diffusion models~\cite{rombach2022LatentDiff,zhang2023addingControlNet} that pre-trained on billions of high-quality images for getting vivid plausible outputs.
In the following paragraphs, we will distinguish between these technologies in detail, and we list comparisons with several representative methods in Tab.~\ref{tab:distinguish}.

\vspace{0.1cm}
\noindent\textbf{OG-NeRF.}
The original NeRF technology~\cite{NeRF_mildenhall2020nerf} overfits the specific scene with tens or hundreds of posed input views and requires per-scene optimization. Targeting these issues, generalizable NeRF~\cite{PixelNeRF_yu2021pixelnerf,FE_NVS_guo2022fast,lin2023visionnerf} is proposed which learns the general 3D prior across multiple scenes given very sparse reference images, which can naturally be applied to the O-NVS task.

(i) \textit{Implicit condition.} Early works~\cite{CodeNeRF_jang2021codenerf,ShaRF_rematas2021sharf} employ the implicit condition paradigm, which \textit{implicitly} encode the scene information into the latent code based on the auto-decoder framework and requires tedious test-time optimization to find the latent code for new scenes.

(ii) \textit{Explicit condition.} Another line of works~\cite{PixelNeRF_yu2021pixelnerf,FE_NVS_guo2022fast,lin2023visionnerf} construct the condition \textit{explicitly} with the help of an encoder module which extracts condition features from the reference images, and can generalize to a new scene by a single feed-forward pass. However, they inevitably suffer the blurry issues, especially when the target view is far from the source view, since they rely highly on the condition features from the limited reference image. 

Targeting this issue, we propose GD$^2$-NeRF, a coarse-to-fine framework to compensate for the details using generative models. At the coarse stage, we first inject the GAN model into the existing OG-NeRF pipeline to primarily relieve the blurry issue through learning in-distribution detail priors. Building on top of this, at the fine stage, we further exploit more vivid out-distribution priors from  the pre-trained diffusion model~\cite{rombach2022LatentDiff,zhang2023addingControlNet}.

Recent work~\cite{gu2023nerfdiff} also tries to relieve the blurry issue by adding a generative model. However, it co-trains a diffusion model with a conditional NeRF from scratch and needs tedious inference-time finetuning for each reference image. While our framework is finetuning-free and the pre-trained diffusion model~\cite{rombach2022LatentDiff,zhang2023addingControlNet} on the large-scale high-resolution dataset is directly employed.


\vspace{0.1cm}
\noindent\textbf{Geometry-free Methods.} Several works~\cite{srt22,watson20223DiM} also attempt to train a conditional model on posed images while without modeling the underlying geometry. For example, 3DiM~\cite{watson20223DiM} trains a pose-conditioned diffusion model on pairs of posed images, and inference in an auto-regression manner. However,  though better at per-image quality (high FID), it fails to show smooth 3D consistency than the OG-NeRF methods due to the lack of geometry constraints.

\begin{table}[t]
\setlength{\abovecaptionskip}{0cm}
\footnotesize
     \caption{\textbf{Distinguish between existing representative technologies that can solve the O-NVS task under different settings (\S~\ref{sec:related_work}).} }
    \label{tab:distinguish}
    \setlength\tabcolsep{3.6pt}

\begin{tabular}{c|l|ccc}
    \rowcolor[gray]{.9}
\hline
\multicolumn{1}{c|}{\textbf{Technology}  }                                                                    & \textbf{Method} & \textbf{\begin{tabular}[c]{@{}c@{}}Inference-time \\ Finetuning-free\end{tabular}}  & \textbf{\begin{tabular}[c]{@{}c@{}}GAN \\ Model \end{tabular}} & \textbf{\begin{tabular}[c]{@{}c@{}}Diffusion \\ Model \end{tabular}}  \\ \hline \hline
\multirow{2}{*}{3D GAN}                                               & EG3D-PTI~\cite{chan2022EG3D}             & \xmark           & \cmark      & \xmark      \\
                                                                      & Pix2NeRF~\cite{Pix2NeRF_cai2022pix2nerf} & \cmark           &\cmark       & \xmark      \\ \hline
\multirow{1}{*}{Geometry-free}                                        & 3DiM~\cite{watson20223DiM}               & \cmark           & \xmark      & \cmark      \\ \hline
                                                                                          
\multirow{4}{*}{\begin{tabular}[c]{@{}c@{}}Image-to-3D\end{tabular}}  & DietNeRF~\cite{Jain_2021_ICCV_dietnerf}  & \xmark           & \xmark      & \xmark     \\
                                                                      & SinNeRF~\cite{Xu_2022_SinNeRF}           & \xmark           & \xmark      & \xmark     \\
                                                                      & Zero-123-NVS~\cite{liu2023zero123}           & \cmark          & \xmark      & \cmark     \\
                                                                      & Zero-123~\cite{liu2023zero123}           & \xmark           & \xmark      & \cmark     \\
                                                                      & Make-it-3D~\cite{tang2023makeit3d}       & \xmark           & \xmark      & \cmark     \\ \hline
\multirow{2}{*}{OG-NeRF}                                              & NeRFDiff~\cite{gu2023nerfdiff}           & \xmark           & \xmark      & \cmark     \\
                                                                      & \textbf{Ours}                            & \cmark           & \cmark      & \cmark     \\ \hline
\end{tabular}

\end{table}

\vspace{0.1cm}
\noindent\textbf{3D GAN.} 
In recent years, with the success of GANs in 2D image synthesizing~\cite{GAN_resolution_brock2018large,GAN_resolution_choi2018stargan,GAN_resolution_karras2019style} and the impressive performance of Neural Radiance Fields (NeRF)~\cite{NeRF_mildenhall2020nerf}, a bunch of works~\cite{GRAF_schwarz2020graf,PiGAN_chan2021pi,GIRAFFE_niemeyer2021giraffe,chan2022EG3D} include NeRF into GAN models as inductive bias for 3D-aware image synthesis. Typically, they can solve the O-NVS task via \textit{GAN inversion} technology~\cite{chan2022EG3D}. However, there mainly exist the following drawbacks: 

(i) \textit{Per-scene optimization.} It requires tedious optimization to find the corresponding latent code for each reference image. Though~\cite{Pix2NeRF_cai2022pix2nerf}further includes the encoder model into the existing framework~\cite{PiGAN_chan2021pi} to obtain a conditional 3D GAN model, the performance is still unsatisfactory since it is trained on a collection of unposed images, \ie, unsupervised.

(ii) \textit{Specific category.} Similar to the traditional GAN methods, they usually only work on a specific category like cat, car, etc.

In contrast to these methods~\cite{GRAF_schwarz2020graf,PiGAN_chan2021pi,GIRAFFE_niemeyer2021giraffe,chan2022EG3D}, our method does not require per-scene optimization given reference images and can generalize across different categories, even for real-world complex scenes.

\vspace{0.1cm}
\noindent\textbf{Large-model-based Image-to-3D.} 
As the recent advances on pre-trained large models~\cite{caron2021emerging,radford2021CLIP,rombach2022LatentDiff,zhang2023addingControlNet}, a bunch of works~\cite{Jain_2021_ICCV_dietnerf,Xu_2022_SinNeRF,liu2023zero123,tang2023makeit3d} explore using them to solve the O-NVS task in a per-scene optimization manner for getting plausible 3D representations.

(i) \textit{DINO\&CLIP-based.} Early attempts DietNeRF~\cite{Jain_2021_ICCV_dietnerf} and SinNeRF~\cite{Xu_2022_SinNeRF} use DINO~\cite{caron2021emerging} or CLIP~\cite{radford2021CLIP} to constrain the distance between the rendered novel views and the reference view in the feature space. However, the feature space constraint struggles to provide fine-grained information, and ~\cite{Xu_2022_SinNeRF} only works in nearby views.

(ii) \textit{Diffusion-based.} Recently, several works~\cite{deng2023nerdi,xu2022neurallift,liu2023zero123,tang2023makeit3d} attempt to lift the fine-grained 2D generative prior in pre-trained diffusion models~\cite{rombach2022LatentDiff,zhang2023addingControlNet} to plausible 3D representations. For instance, Zero123~\cite{liu2023zero123} first finetunes the latent diffusion~\cite{rombach2022LatentDiff} on a synthetic dataset~\cite{deitke2023objaverse} to inject the viewpoint condition (Zero123-NVS). Though Zero123-NVS can work in an inference-time finetuning-free manner, it achieves poor 3D consistency considering the undeterministic nature of diffusion models. Therefore, it further optimizes a 3D representation with the finetuned diffusion model using SJC~\cite{wang2023SJC} framework. 

Different from all the methods above~\cite{deng2023nerdi,xu2022neurallift,Jain_2021_ICCV_dietnerf,Xu_2022_SinNeRF,liu2023zero123,tang2023makeit3d}, our framework requires NO per-scene optimization, and the pre-trained diffusion model is also fixed with no further finetuning in our fine-stage method 3DE.

\subsection{\textbf{Diffusion-based Video Editing}}

With the success of text-to-image diffusion models~\cite{rombach2022LatentDiff,zhang2023addingControlNet}, recent works~\cite{wu2023tune-a-video,text2video-zero,ceylan2023pix2video,qi2023fatezero,geyer2023tokenflow,yang2023rerender} try to solve the video editing task via the pre-trained text-to-image diffusion models, either with finetuning~\cite{wu2023tune-a-video} or in a zero-shot manner~\cite{text2video-zero,qi2023fatezero,ceylan2023pix2video,geyer2023tokenflow,yang2023rerender}. The main challenge for extending an image diffusion model to a video editing task is to ensure the consistency between video frames. Early works~\cite{wu2023tune-a-video,text2video-zero,ceylan2023pix2video} mainly ensure the global appearance consistency by inflating the self-attention module to the temporal cross-attention module. Besides, ~\cite{yang2023rerender} further includes the optical flow as the correspondence constraint, while~\cite{geyer2023tokenflow} directly uses the correspondences calculated during DDIM inversion. However, all of them take consecutive video sequences with similar contents as inputs and can not process an arbitrary view in a 3D manner. 

Differently, in this work, our Diff3DE makes the early attempts to extend the existing methods to a 3D-aware detail enhancer that maintains 3D consistency between different views that contain large content variance (\eg, front and back views of a car) and supports the process of an arbitrary view. 

\section{\textbf{Preliminary}}
\noindent\textbf{Overview.} In this section, we first briefly introduce the Neural Radiance Fields (NeRF)~\cite{NeRF_mildenhall2020nerf} in \S~\ref{sec:nerf} and its one-shot generalizable variant~\cite{PixelNeRF_yu2021pixelnerf} in \S~\ref{sec:ognerf}. Then, we analysis the limitation of esisting OG-NeRF in \S~\ref{sec:ognerf_limit}. Finally, we introduce the GAN model in \S~\ref{sec:gan_model} and diffusion model in \S~\ref{sec:diff_model}.

\vspace{-2mm}
\subsection{\textbf{Neural Radiance Fields}}
\label{sec:nerf}
NeRF parameterizes the 3D volume space as a continuous implicit function $f_{nerf}(\cdot)$ represented by a neural network, \eg, multi-layer perceptron (MLP). Given a target view with pose $\textbf{P}_t$, it queries the 3D volume space via marching a ray $\textbf{r}(z) = \textbf{o} + z \textbf{d}$ for each pixel on its image plane, where $\textbf{o} \in  \mathbb{R}^3$ represents the camera center, $\textbf{d} \in  \mathbb{R}^3 $ represents the ray unit vector, and $z \in  \mathbb{R}^1$ is the depth between a pre-defined bounds $[z_n, z_f]$. Then, for each 3D point $\textbf{x} \in  \mathbb{R}^3 $ on a marched ray, its density $\sigma \in  \mathbb{R}^1$ and RGB color $\textbf{c} \in  \mathbb{R}^3$ is predicted by:
\begin{equation}
    (\sigma, \textbf{c}) = f_{nerf }(\textbf{x}, \textbf{d}).
\end{equation}
After that, the predicted RGB color $\textbf{c}$ along a ray $\textbf{r}$ is accumulated via the differentiable volumetric rendering operation: 
\begin{equation}
    \label{accumulate}
    \hat{\textbf{C}}(\textbf{r}) = \int_{z_n}^{z_f} T(z) \sigma (z) \textbf{c}(z) dz,
\end{equation}
where $T(z) = exp (- \int_{z_n}^z \sigma(s) ds )$ represents the probability that the ray travels from $z$ to $z_n$. 
NeRF employs a hierarchical coarse-to-fine strategy for sampling discrete 3D points along rays and then approximates the integral using numerical quadrature~\cite{max1995optical}. 
Finally, for each pixel, the accumulated RGB color $\hat{\textbf{C}}(\textbf{r})$ is supervised by the ground truth color $\textbf{C}(\textbf{r})$ through the mean squared error:
 
\begin{equation}
    \label{eq_photometricloss}
    \mathcal{L}_{NeRF} = \frac{1}{| \mathcal{R}(\textbf{P}_t)|} \sum_{\textbf{r} \in \mathcal{R}(\textbf{P}_t)} ||\hat{\textbf{C}}(\textbf{r}) - \textbf{C}(\textbf{r})||_{2}^2 , 
\end{equation}
where $\mathcal{R}(\textbf{P}_t)$ is the set of all marched rays from target pose $\textbf{P}_t$.

\vspace{-2mm}

\subsection{\textbf{One-shot Generalizable NeRF}}
\label{sec:ognerf}
NeRF is optimized per scene and requires tens or hundreds of posed input views and a time-consuming optimization process to memorize the scene. To relieve this issue and realize O-NVS, One-shot Generalizable Neural Radiance Fields (OG-NeRF)~\cite{PixelNeRF_yu2021pixelnerf} is proposed to learn the general 3D prior across multiple scenes. At the core of OG-NeRF is to take an additional condition feature extracted from the reference image $\textbf{I}_s$ as input. A typically used method for getting the condition feature~\cite{PixelNeRF_yu2021pixelnerf,lin2023visionnerf} is to project the query point $\textbf{x}$ back to the feature map of the reference image, and fetch a feature via interpolation:

\begin{equation}
\label{eq:ognerf}
    (\sigma, \textbf{c}) = f_{ognerf}(\textbf{x}, \textbf{d}, W(\pi(\textbf{x}))),
\end{equation}
where $W=E({\textbf{I}_s})$ is the extracted feature map via encoder $E$, and $\pi(\textbf{x})$ indicates the projection of $\textbf{x}$ on the reference image plane. Note that, for better generalization ability, $\textbf{x}$ and $\textbf{d}$ here are under the coordinate of the reference view, \ie, the relative one, instead of the world coordinate. 


\vspace{-2mm}
\subsection{\textbf{Limitations of Existing OG-NeRF Methods}}
\label{sec:ognerf_limit}
The main drawback of such an encoder-only paradigm is the high reliance on the reference image, which 
may include misleading information, especially for the projection-based ones~\cite{PixelNeRF_yu2021pixelnerf,lin2023visionnerf} (see the upper part of Fig. \ref{fig_idea}).
To address this issue, 
we propose the coarse-to-fine generative detail compensation perspective. Specifically, at the coarse stage, we first propose the OPP that efficiently insert a GAN model into the existing OG-NeRF pipeline for capturing the in-distribution object details. Then, at the fine stage, with Diff3DE we further leverage the out-distribution detail priors from the pre-trained diffusion model~\cite{rombach2022LatentDiff,zhang2023addingControlNet} in a 3D-consistency manner for more vivid outputs with plausible details.

\subsection{\textbf{GAN Model}}
\label{sec:gan_model}

We inject the GAN model to the OG-NeRF pipeline at the coarse stage for capturing primary in-distribution detail priors from the training dataset. For the training of the GAN model, we employ the commonly used  non-saturating GAN objective~\cite{GAN_goodfellow2014generative} with $R_1$ gradient penalty~\cite{R1reg_mescheder2018training}:
\begin{equation}
\small
\begin{split}
    & \mathcal{L}_{GAN} = 
    \mathbb{E}_{\textbf{I}_{in} \sim p_{in}} [ D(G(  \textbf{I}_{in} )) ]   +  \mathbb{E}_{\textbf{I}_{real} \sim p_{real}} [-D(\textbf{I}_{real})], 
\end{split}
\end{equation}
 where $p_{in}$ represents the distribution of the generator input, $p_{real}$ indicates the distribution of the training data, and the formulation of $R1$ penalty  is omitted here for simplicity.

 Except for the GAN objective, we also employ the perceptual loss $\mathcal{L}_{PER}$~\cite{Perceptual_wu2020unsupervised}   and MSE loss $\mathcal{L}_{MSE}$~\cite{GIRAFFE_niemeyer2021giraffe} for the reconstruction from $G(\textbf{I}_{in})$ to its corresponding ground truth $\textbf{I}_{real}$. Therefore, the overall objective  $\mathcal{L}_{G}$ for the GAN training is:
 \begin{equation}
 \label{eq:generative_objectives}
     \mathcal{L}_{G} = \lambda_{GAN} \mathcal{L}_{GAN} +\lambda_{PER} \mathcal{L}_{PER} + \mathcal{L}_{MSE},
 \end{equation}
 where $\lambda_{GAN}$ and $\lambda_{PER}$ are the weights for GAN loss and perceptual loss, separately.

\subsection{\textbf{Diffusion Model}}
\label{sec:diff_model}

Diffusion probabilistic model has been broadly researched recently~\cite{croitoru2023diffusion0,dhariwal2021diffusion1,rombach2022LatentDiff,zhang2023addingControlNet} due to its strong generative power. It approximates the data distribution via progressively removing noises from a Gaussian \textit{i.i.d} noised image. Stable Diffusion~\cite{rombach2022LatentDiff} makes the early attempt to operate on the latent space of a pre-trained auto-encoder to efficiently get high-resolution results. Building on this, ControlNet~\cite{zhang2023addingControlNet} enables more accurate and flexible controls via incorporating additional condition signals with a residual architecture, \eg, edge, pose, etc. 

\vspace{0.1cm}
\noindent\textbf{DDIM Inversion.} DDIM is a deterministic denoising-stage sampling algorithm proposed by~\cite{song2020DDIM}. It can be used in a reversed order to find the corresponding noises of an image in a \textit{non-optimization manner}, which is commonly used in image/video editing tasks~\cite{yang2023rerender,geyer2023tokenflow,Tumanyan_2023_CVPR_PnP} for primarily maintaining the original contents of input images.

\vspace{0.1cm}
\noindent\textbf{Inflated Self-attention.} Usually, a U-Net $\epsilon_{\theta}$ with attention blocks~\cite{rombach2022LatentDiff,zhang2023addingControlNet} is employed for predicting the noise. To improve the temporal consistency when processing a sequence of frames, recent methods~\cite{wu2023tune-a-video,yang2023rerender,geyer2023tokenflow} inflate the original self-attention to incorporate information from other frames. In detail, given the sequence of projected query features $\{\textbf{Q}_i\}_{i=1}^{k}$, key features $\{\textbf{K}_i\}_{i=1}^{k}$, and value features $\{\textbf{V}_i\}_{i=1}^{k}$, the Inflated Self-Attention (ISA) is calculated as follows: 
\begin{equation}
    \phi_{i} = Softmax(\frac{\textbf{Q}_i[\textbf{K}_1, ...,\textbf{K}_k]^T}{\sqrt{d}}) [\textbf{V}_1, ..., \textbf{V}_k],
\end{equation}
where $\phi_{i}$ is the ISA output tokens of the $i$-th frame.

\section{\textbf{Coarse Stage: From Tandem to Parallel Pipelines}}
\label{sec:coarse_stage_opp}

\begin{figure}[t]
\setlength{\abovecaptionskip}{0cm}
\small
\centering
\includegraphics[width=1.0\linewidth]{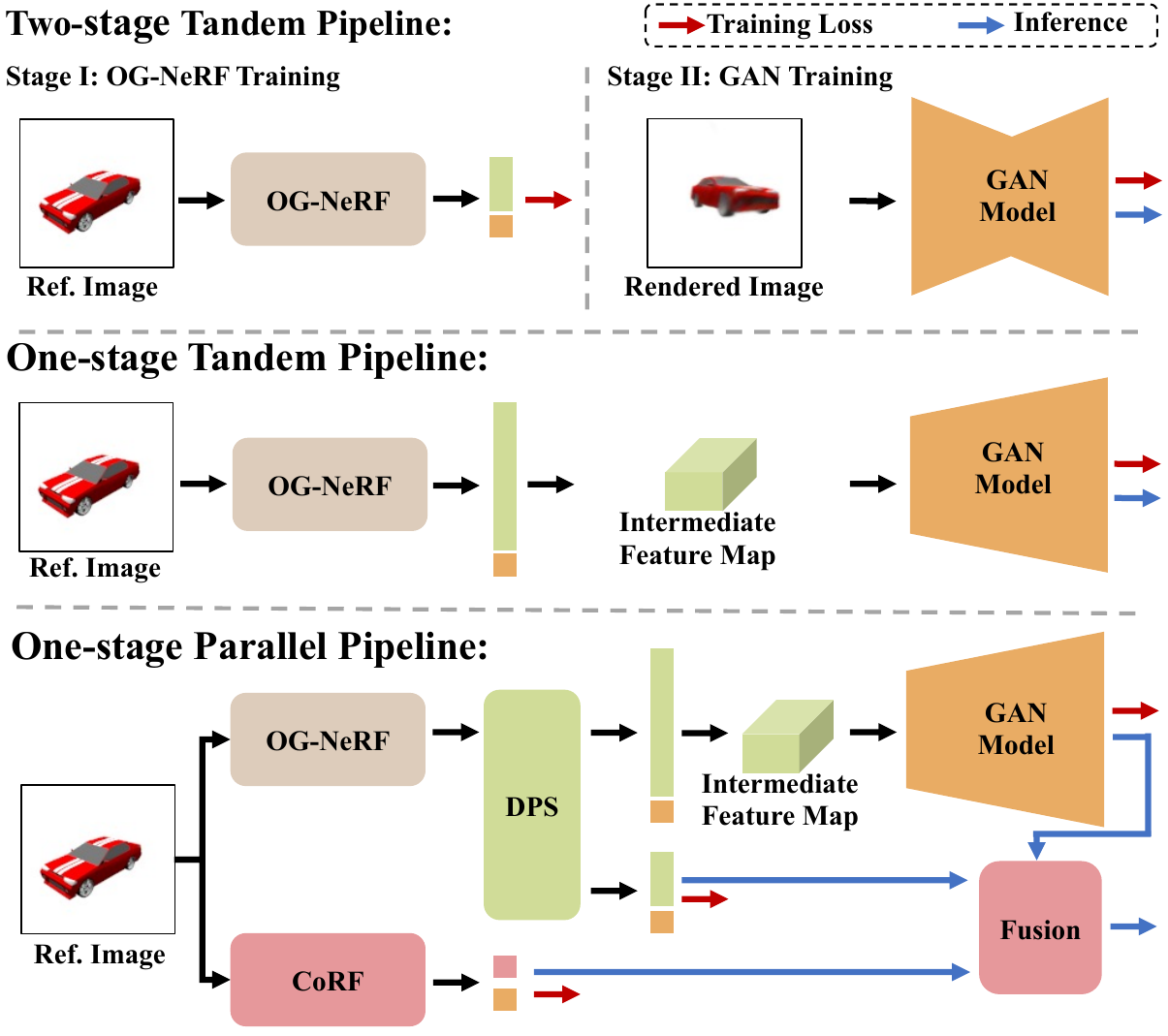}
\caption{\textbf{Overview of two basic tandem pipelines (\S~\ref{sec:ttp}, \S~\ref{sec:otp}) and our proposed One-stage Parallel Pipeline (OPP, \S~\ref{sec:opp})} for integrating the GAN model into the OG-NeRF framework at the COARSE STAGE. }

\label{fig_simple_pipelines}
\end{figure}


\begin{figure*}[t]
\setlength{\abovecaptionskip}{0cm}
\small
\centering
\includegraphics[width=0.99\linewidth]{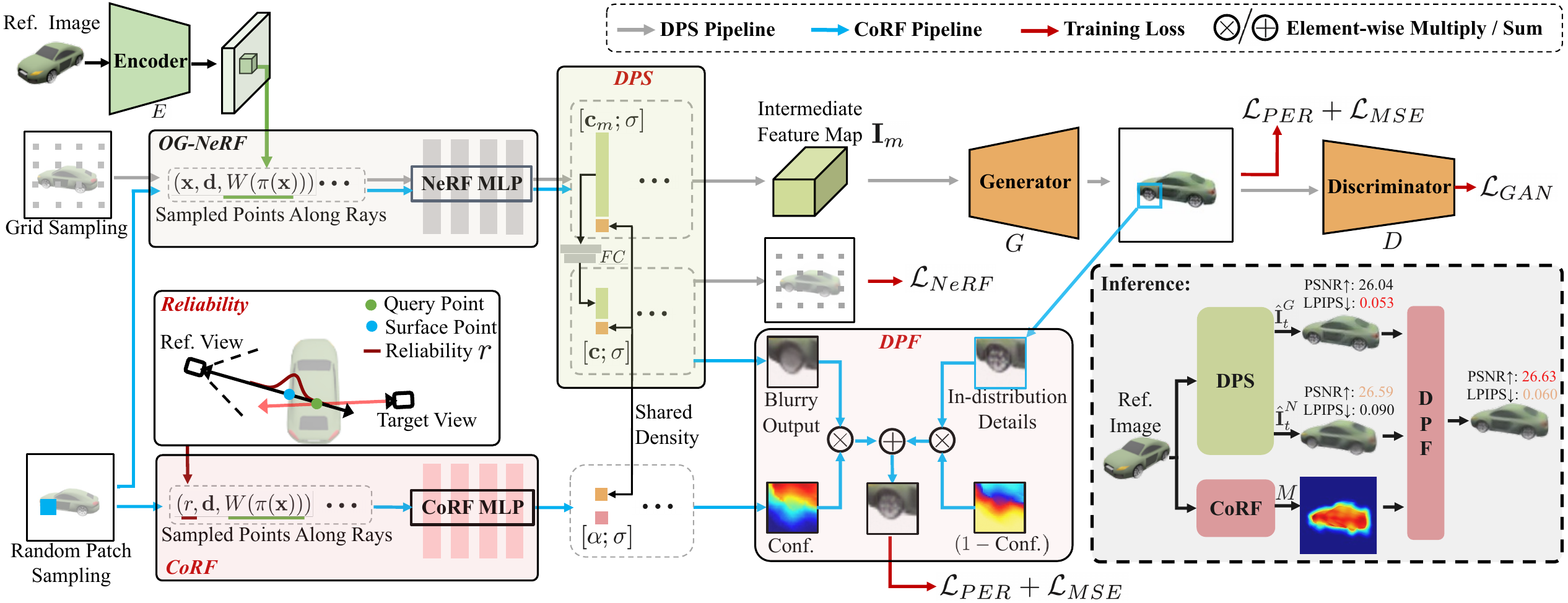}
\caption{\textbf{Overview of the COARSE-STAGE method OPP for including in-distribution details from the training data (\S~\ref{sec:opp}).}  It is built on the one-stage tandem pipeline (the first row) and efficiently integrates the GAN and OG-NeRF models in a unified parallel framework with DPS, CoRF, and DPF.}
\label{fig_pipelines}
\vspace{-2mm}
\end{figure*}

\noindent\textbf{Overview.} This section introduces the coarse stage method that injects the GAN model into the OG-NeRF pipeline. We first briefly analyze two basic tandem pipelines that directly build the GAN model on top of the OG-NeRF model (\S~\ref{sec:ttp}, \S~\ref{sec:otp}), as shown by Fig.~\ref{fig_simple_pipelines}. Then, in \S~\ref{sec:opp}, we further present a stronger One-stage Parallel Pipeline (OPP)  that solves the contradiction between sharpness and fidelity efficiently. The pipeline of OPP is illustrated in Fig. \ref{fig_pipelines}.

\subsection{\textbf{Two-stage Tandem Pipeline}}
\label{sec:ttp}
Different from OG-NeRF which supervises at the pixel level, the GAN model requires the input to be semantically complete at the image level, ideally, the full-sized image. However, rendering out the full-sized image during training is intractable considering the memory-intensive rendering process. A naive solution is to independently train the OG-NeRF at the first stage. Then, in the second stage, the full-sized novel view images are rendered with the previously trained OG-NeRF (fixed and inference only) pixel by pixel, and a GAN model is trained with these prepared full-sized images and their corresponding ground truth images, as illustrated by the first row in Fig.~\ref{fig_simple_pipelines}. 

The main drawbacks of such a pipeline are two-fold: (i) Though the sharpness (\ie, details) can be improved with such a pipeline, the fidelity of the generated details can hardly be guaranteed. The independently trained GAN model is easily biased towards the sharpness details without the joint optimization with the OG-NeRF model which constrains the fidelity. (ii) The two-stage training procedure is tedious. 

\vspace{-2mm}
\subsection{\textbf{One-stage Tandem Pipeline}}
\label{sec:otp}
To relieve the memory cost and facilitate the end-to-end training, the key is to reduce the rendered output size from $H \times W$ to $H_m \times W_m$. However, directly reducing the output size will inevitably cause the loss of 3D information from the volume space. To compensate for it, a feasible solution is to simultaneously improve the output dimension from $\textbf{c} \in \mathbb{R}^3$ to a higher value $\textbf{c}_m \in \mathbb{R}^{h_m}, h_m > 3$, as in~\cite{GIRAFFE_niemeyer2021giraffe} and illustrated by the second row of Fig.~\ref{fig_simple_pipelines}. Then, similar as Eq. \ref{accumulate}, $\textbf{c}_m$ is accumulated along the ray $\textbf{r}$ as follows:
\begin{equation}
\label{Cm_accumulate}
    \hat{\textbf{C}}_m(\textbf{r}) = \int_{z_n}^{z_f} T(z) \sigma (z) \textbf{c}_m(z) dz.
\end{equation}
Notably, for the sampling of rays, different from the pixel-wise supervision method \cite{PixelNeRF_yu2021pixelnerf} which randomly samples rays among all target poses around the object, we employ the \textit{grid sampling strategy} which first randomly samples a target pose, and then split its corresponding image plane of size $H \times W$ into $H_m \times W_m$ grids. After that, the rays tracing through the grid centers are sampled for accumulating with Eq. \ref{Cm_accumulate} and get the intermediate low-resolution yet high-dimensional feature map $\hat{\textbf{I}}_m \in \mathbb{R}^{H_m \times W_m \times h_m}$ via reshaping. 
Finally, $\hat{\textbf{I}}_m$ is sent to a light-weight up-sampling model $G(\cdot)$ for the full-sized output $ \hat{\textbf{I}}_t^{G} \in \mathbb{R}^ {H \times W \times 3}$, which is finally supervised with a discriminator $D(\cdot)$. 

Compared with the naive two-stage tandem pipeline, this end-to-end fashion makes the GAN model benefits more fidelity from the OG-NeRF model which extracts 3D information directly from the volume space. Therefore, it performs better than the two-stage tandem pipeline, especially at the fidelity metrics, \eg, PSNR, SSIM. 

\vspace{-2mm}
\subsection{\textbf{One-stage Parallel Pipeline}}
\label{sec:opp}
After including the GAN model as a tandem pipeline, the \textit{sharpness} (\eg, LPIPS, FID) is significantly improved since the GAN model can capture the prior object details from the abundant training data. However, we find it hard to maintain the \textit{fidelity} (\eg, PSNR, SSIM) as well as the independently trained OG-NeRF model, even with the more coherent one-stage tandem pipeline.

To tackle such contradiction between two paradigms, we further present the simple yet effective One-stage Parallel Pipeline (OPP) that is built on the one-stage tandem pipeline with the proposed Dual-Paradigm Structure (DPS), Confidence Radiance Fields (CoRF), and Dual-Paradigm Fusion (DPF), as illustrated in Fig. \ref{fig_pipelines}. 
DPS makes it possible for these two paradigms to be optimized in parallel within a single framework, and CoRF further learns a confidence map that can adaptively give the blurry part with lower confidence. Finally, DPF integrates the outputs from two paradigms effectively with the learned confidence map. 

\vspace{0.1cm}
\noindent\textbf{Dual-Paradigm Structure.}
 In the one-stage tandem pipeline,  the output of the OG-NeRF MLP is represented as $[\textbf{c}_m;\sigma] \in \mathbb{R}^{h_m + 1}$, where $[;]$ indicates the concatenation operation, $\textbf{c}_m \in \mathbb{R}^{h_m} $ is the high-dimensional hidden color, and $\sigma \in \mathbb{R}^{1}$ is the density. Then, we send $\textbf{c}_m$ to an additional fully connected layer $FC(\cdot): \mathbb{R}^{h_m} \rightarrow \mathbb{R}^{3} $ as follows:
\begin{equation}
    \textbf{c} = FC(\textbf{c}_m),
\end{equation}
where $\textbf{c} \in \mathbb{R}^{3}$ is the RGB color. After that, $\textbf{c}$ and $\textbf{c}_m$ are accumulated by Eq. \ref{accumulate} and Eq. \ref{Cm_accumulate} with the shared density $\sigma$ and get $\hat{\textbf{C}} \in \mathbb{R}^{3}$ and $\hat{\textbf{C}}_m \in \mathbb{R}^{h_m}$, separately. Finally, $\hat{\textbf{C}}$ is supervised with the ground truth pixel color with Eq. \ref{eq_photometricloss}, and $\hat{\textbf{C}}_m$ marched from the same target pose formulates the intermediate feature map $\hat{\textbf{I}}_m$, which is sent to the up-sampling module for the GAN supervision.

With DPS, the OG-NeRF and GAN model can be optimized in parallel with a single training process. Intuitively, $\textbf{c}_m$ can be seen as the mapping of RGB color $\textbf{c}$ in a high-dimensional feature space. We inverse it back to the RGB space with a lightweight fully connected layer for the NeRF-style supervision.
The shared density also profits the communications between two paradigms.

\vspace{0.1cm}
\noindent\textbf{Confidence Radiance Fields.}
An ideal solution to integrate the two paradigms is to automatically detect the blurry part on the OG-NeRF output and then complement the corresponding details from the GAN output. Motivated by this, we propose the novel Confidence Radiance Fields (CoRF) that gives each pixel a confidence score reflecting the degree of clarity.

Specifically, for each sampled point $\textbf{x}$, 
we assume that the reliability of
the projected condition feature 
is determined by the distance between $\textbf{x}$ and the projected surface point $\textbf{s}$ since it reflects the occlusion information, as shown in Fig.~\ref{fig_pipelines}. Therefore, we define reliability as
    $r = \mathcal{G}(z_{\textbf{x}} - z_{\textbf{s}})$,
where $\mathcal{G}$ is a gaussian function, and $z_{\textbf{x}}$, $z_{\textbf{s}}$ are the depth of $\textbf{x}$ and $\textbf{s}$ from the reference view, respectively. Notably, $z_{\textbf{x}}$ can be achieved by simply using the pose information of the reference view,  while for $z_{\textbf{s}}$, we first render a coarse depth map (\eg, $16\times16$) from the reference view and resize it to the full size (\eg, $128\times128$), then index the depth via projection.

In a nutshell, the final CoRF is represented as:
\begin{equation}
    \alpha  =  f_{corf}(r, \textbf{d}, W(\pi(\textbf{x})),
\end{equation}
where $\alpha \in [0,1]$ represents the confidence score. Then, similar to Eq.~\ref{accumulate}, the confidence score along a ray $\textbf{r}$ is accumulated with the differentiable volumetric rendering using the same density value as in DPS, and get the final confidence score $\hat{\alpha}(\textbf{r}) \in \mathbb{R}^1$.

\vspace{0.1cm}
\noindent\textbf{Dual-Paradigm Fusion.}
After that, for each ray $\textbf{r}$, we fuse the RGB values from two paradigms with:
\begin{equation}
    \hat{\textbf{C}}^{'}(\textbf{r}) = \hat{\textbf{C}}(\textbf{r}) * \hat{\alpha}(\textbf{r})  +  \hat{\textbf{I}}_t^{G}(\textbf{r}) * (1 - \hat{\alpha}(\textbf{r}) ), 
\end{equation}
where $\hat{\textbf{C}}(\textbf{r})$, $\hat{\textbf{I}}_t^{G}(\textbf{r}) \in \mathbb{R}^3$ are the RGB values from OG-NeRF and GAN model, respectively, and $\hat{\textbf{C}}^{'}(\textbf{r})$ is the final fused one where we conduct CoRF training objectives on.


\vspace{0.1cm}
\noindent\textbf{Training \& Inference.} During the training stage, since we expect the DPF output to have both high sharpness and fidelity, except for the \textit{grid sampling} used for DPS training, we additionally employ a \textit{random patch sampling strategy} for the CoRF learning, which can output a semantic patch that supports the perceptual supervision, as illustrated by Fig.~\ref{fig_pipelines}. The overall loss for the training of OPP is:
\begin{equation}
    \mathcal{L}_{OPP} = \mathcal{L}_{G} + \mathcal{L}_{NeRF} + \mathcal{L}_{CoRF} ,
\end{equation}
where $\mathcal{L}_{CoRF} = \mathcal{L}_{PER} + \mathcal{L}_{MSE}$. Considering the training stability, we empirically first train the two paradigms till convergence and then finetune the CoRF for several epochs with the two paradigms frozen. 

During the inference stage of coarse-stage OPP, we have $\hat{\textbf{I}}_t^{G}$ from the GAN model via grid sampling, $\hat{\textbf{I}}_t^{N}$ from the OG-NeRF paradigm and the confidence map $M$ from the CoRF with the full-sized sampling. Then, they are aggregated via DPF for the final output, as shown in Fig. \ref{fig_pipelines}. We emphasize that the rendering of grid sampling (\eg, $16\times16$) is quite efficient (over $40$ times faster) compared with the full-sized rendering (\eg, $128\times128$), and the CoRF MLP is rather lightweight ($0.09M$), therefore the additional computational cost compared with the original OG-NeRF is rather small, which will be discussed in detail in the appendix. 

\section{\textbf{Fine Stage: Diffusion-based 3D Enhancer}}
\label{fine_stage_Diff3DE}

\begin{figure*}[t]
\setlength{\abovecaptionskip}{0cm}
\small
\centering
\includegraphics[width=1.0\linewidth]{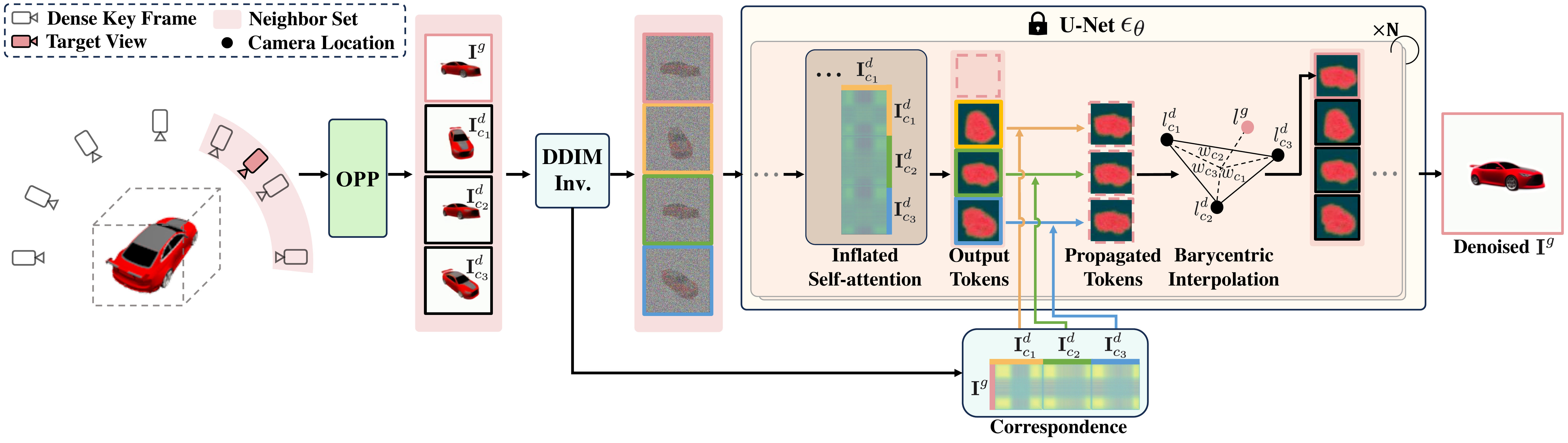}
\vspace{-4mm}
\caption{\textbf{Overview of the FINE-STAGE method Diff3DE for including out-distribution details from the pre-trained diffusion model~\cite{rombach2022LatentDiff,zhang2023addingControlNet} (\S~\ref{fine_stage_Diff3DE}).} We first fix $N_k$ dense keyframes around the dome. Then, for each target view, we select $3$ neighbor keyframes based on the cosine similarity. For each diffusion time step and attention block, the output tokens of the target view are the barycentric interpolation of the propagated tokens from neighbor keyframes, using the correspondence calculated during DDIM inversion. The global 3D consistency is primarily achieved by the 3D-consistent constraint from OPP and further approximated by enforcing the local consistency for each neighbor area.}
\label{fig_diff3de}
\end{figure*}
\noindent\textbf{Motivation.} With OPP, the coarse in-distribution details have been primarily compensated. However, considering the limited size and quality of the training dataset, the results are still unsatisfying in vivid detail. Thus, to break through such limitation, we turn to the rich out-distribution priors from the diffusion models~\cite{rombach2022LatentDiff,zhang2023addingControlNet} pre-trained on billions of high-quality images. 

A naive solution of using such an image diffusion model is to perform super-resolution~\cite{rombach2022LatentDiff,zhang2023addingControlNet} on each rendered view from the OPP individually. Nevertheless, due to the lack of 3D constraints, this will lead to significant inconsistency between different views. Although the recent zero-shot video editing method~\cite{geyer2023tokenflow} has explored maintaining the temporal consistency between edited video frames, it simply assumes the input video to contain very similar contents, and 
directly employing it in the 3D NVS task will lead to the following issues:

(i) \textit{Dispersed sparse keyframes lead to blurry outputs.} At the core of ~\cite{geyer2023tokenflow} is to first maintain the consistency between several keyframes (\eg, 5) using Inflated Self-Attention (ISA)~\cite{wu2023tune-a-video} and then propagate the U-Net self-attention output tokens of keyframes to nearby ones. A possible naive adaptation is to uniformly disperse the keyframes around the dome, however, due to the large variance of contents from dispersed sparse views, the ISA tends to get blurry outputs, as illustrated by Fig.~\ref{fig_ablation_Nk}. On the other hand, simply increasing the keyframe number will greatly increase memory usage, which is unaffordable. 

(ii) \textit{Unable to process arbitrary views.} A 3D enhancer should support the processing of an arbitrary given view. However, as a video processing method, \cite{geyer2023tokenflow} simply assumes the input videos are consecutive and calculates the propagation weights via the frame index, which is not suitable for the arbitrary view processing case.

\vspace{0.1cm}
\noindent\textbf{Diffusion-based 3D Enhancer.} Targeting the aforementioned issues, we propose the fine-stage method Diffusion-based 3D Enhancer (Diff3DE), a 3D extension of ~\cite{geyer2023tokenflow}, as illustrated in Fig.~\ref{fig_diff3de}.  The main idea of Diff3DE is to relax the input of the original ISA from all the keyframes to \textit{neighbor keyframe sets}  selected based on \textit{view distance}.

In detail, we first fix $N_k$ dense keyframes $\{\textbf{I}_i^{d} \}_{i=1}^{N_k}$ uniformly dispersed around the dome. Then, given a target view $\textbf{I}^{g}$, we select its $3$ neighbors $\{ \textbf{I}_{c_1}^{d}, \textbf{I}_{c_2}^{d}, \textbf{I}_{c_3}^{d}\}$ from $\{ \textbf{I}_i^{d}\}$ using the cosine similarity, which can be calculated by the provided camera poses. After that, the neighbor set $\{ \textbf{I}_{c_1}^{d}, \textbf{I}_{c_2}^{d}, \textbf{I}_{c_3}^{d}\}$ are taken as the input of the ISA to ensure the local consistency, and for each diffusion time step and U-Net attention block layer, their output tokens of the ISA module $\{ \phi_{c_1}^{d}, \phi_{c_2}^{d}, \phi_{c_3}^{d}\}$  are further propagated to the target view using the correspondence calculated with original input frames during the DDIM inversion stage (see~\cite{geyer2023tokenflow} for more details), formulating the propagated tokens $\{\phi_{c_1}^{d \rightarrow t},\phi_{c_2}^{d \rightarrow t}, \phi_{c_3}^{d \rightarrow t}\}$. Finally, we perform a weighted sum on  the propagated tokens using the barycentric interpolation:
\begin{equation}
\begin{aligned}
   &  w_{c_1}, w_{c_2}, w_{c_3} = Bary(l^{g}, l^{d}_{c_1}, l^{d}_{c_2}, l^{d}_{c_3}), \\
    & \phi^{t} = w_{c_1} * \phi_{c_1}^{d \rightarrow g} + w_{c_2} * \phi_{c_2}^{d \rightarrow g} + w_{c_3} * \phi_{c_3}^{d \rightarrow g}  ,
\end{aligned}
\end{equation}
where $l$ indicates the camera location, $Bary(\cdot,\cdot,\cdot,\cdot)$ is the function for calculating barycentric weights, and $\phi^t$ is the final aggregated ISA output tokens for $\textbf{I}^{g}$.

With Diff3DE, the processing of each view is determined by its pose instead of the frame index as in~\cite{geyer2023tokenflow}, making it work in a 3D manner. Notably, though the global consistency between all the keyframes is not explicitly constrained in Diff3DE, \ie, we do not send all the keyframes to ISA considering the computation cost, it is approximately maintained by: 1) the input frames are from the 3D-consistent OPP, which naturally provides certain 3D constraints, and 2) enforcing the local consistency for each neighbor area approximates the global consistency.

\begin{figure*}[t]
\setlength{\abovecaptionskip}{0cm}
\small
\centering
\includegraphics[width=0.85\linewidth]{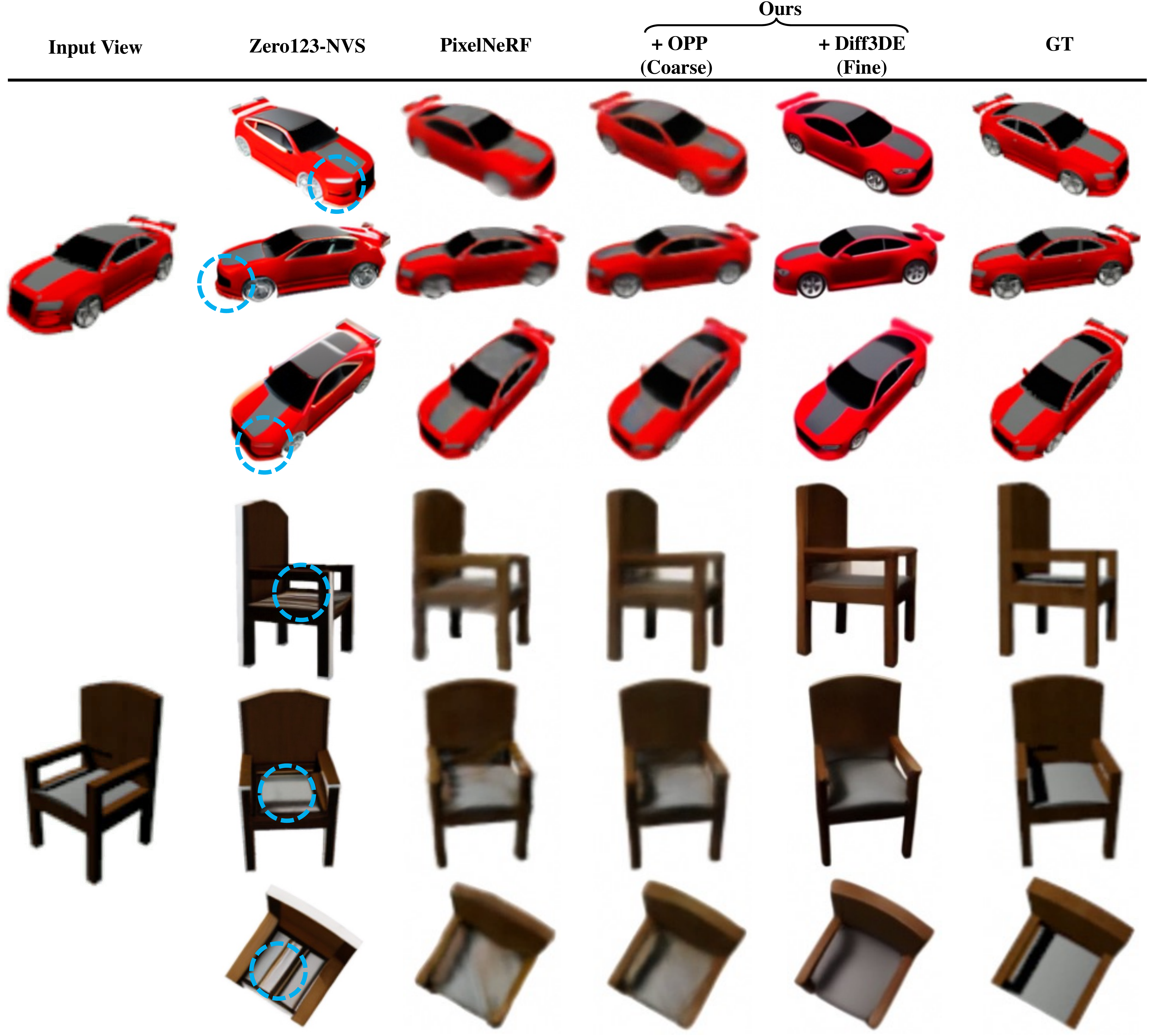}
\caption{\textbf{Qualitative comparisons with  previous methods on ShapeNet Cars \& Chairs (\S~\ref{sec:compari_sota}).} Zero123-NVS~\cite{liu2023zero123} shows significant inconsistency between different views (marked with \textcolor{cyan}{blue circle}), and PixelNeRF~\cite{PixelNeRF_yu2021pixelnerf} gives blurry results. After including our GD$^2$-NeRF framework, the details are greatly improved in a coarse-to-fine manner with decent 3D-consistency. }
\label{fig_srn_nmr_sota}
\vspace{-2mm}
\end{figure*}

\section{\textbf{Experimental Results}}
\subsection{\textbf{Experimental Settings}}

\noindent\textbf{ShapeNet Cars \& Chairs.}
Following previous methods~\cite{PixelNeRF_yu2021pixelnerf, FE_NVS_guo2022fast}, we evaluate on the synthetic large-scale ShapeNet benchmarks~\cite{ShapeNet_chang2015shapenet} with the Cars and Chairs categories following SRN~\cite{SRN_sitzmann2019scene}, which contains $3514$ cars and $6591$ chairs with an image resolution of $128 \times 128$. For each category, we train an individual model.

\noindent\textbf{DTU Dataset.}
For the real-world DTU dataset, unless otherwise specified, we follow the split of ~\cite{PixelNeRF_yu2021pixelnerf} which includes $88$ training scenes and $15$ test scenes. We train at the $128\times128$ resolution and then resize to the original $300 \times 400$ resolution for comparison.

\noindent\textbf{Out-distribution Metrics.}
For the final fine-stage out-distribution method Diff3DE, we evaluate the 3D consistency via rendering a consecutive video around the dome and calculating Pixel-MSE following~\cite{ceylan2023pix2video}, which is the averaged mean-squared pixel error between warped and original frames via optical flow~\cite{teed2020raft}. Notably, since Diff3DE leverages the priors \textit{outside the training dataset distribution}, and mainly focuses on achieving vivid {plausible} details with {3D-consistency}, we do not calculate metrics with the ground truth from the dataset.

\noindent\textbf{In-distribution Metrics.}
For the coarse-stage in-distribution method OPP, we report the commonly used PSNR and SSIM~
\cite{PSNR_wang2004image} as the fidelity metrics, while LPIPS~\cite{LPIPS_zhang2018unreasonable} and FID~\cite{FID_heusel2017gans} as the sharpness metrics.

\vspace{-2mm}
\subsection{\textbf{Implementation Details}}

\noindent\textbf{Architectures.}
(i) \textit{OPP.} We directly take PixelNeRF~\cite{PixelNeRF_yu2021pixelnerf} as the OG-NeRF part of our framework, \ie, ResNet-34 as the encoder and the ResNet-like NeRF MLP. For the verification of different pipelines,  
we employ the commonly used U-Net~\cite{Unet_ronneberger2015u} as the generator for the two-stage pipeline, and the lightweight up-sampling module ($0.11 M$) used in one-stage pipelines is in line with ~\cite{GIRAFFE_niemeyer2021giraffe}.
The discriminator in all the pipelines also shares the same architecture as in~\cite{GIRAFFE_niemeyer2021giraffe}. The CoRF is composed of three fully connected layers ($0.09M$). 

(ii) \textit{Diff3DE.} For Diff3DE, we employ the pre-trained ControlNet-Tile~\cite{zhang2023addingControlNet} as the diffusion model, which conditions on the text together with the input image to perform super-resolution.


\begin{figure*}[t]
\small
\centering
\includegraphics[width=0.85\linewidth]{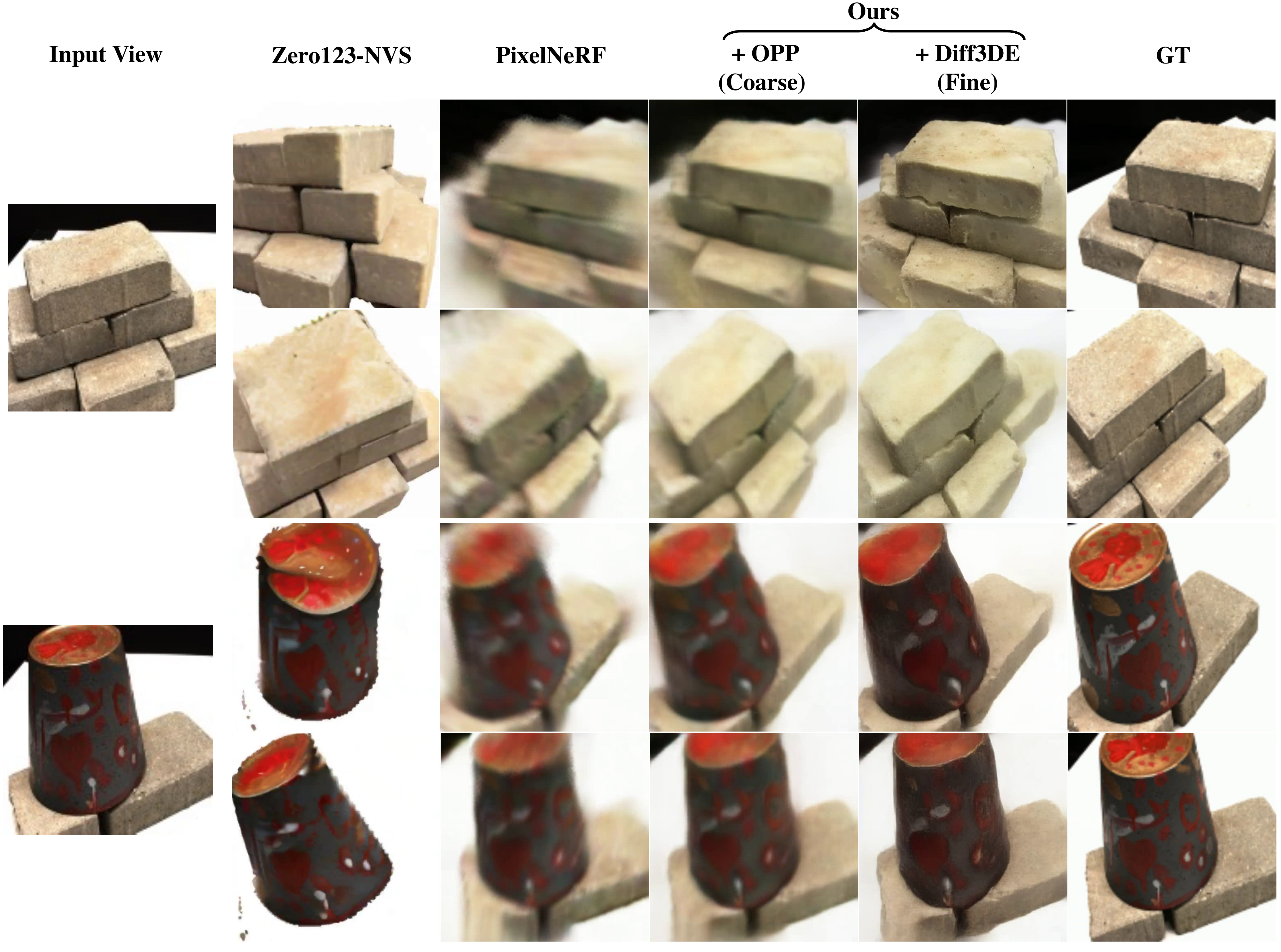}
\vspace{-3mm}
\caption{\textbf{Qualitative comparisons with previous methods on the DTU dataset~\cite{DTU_jensen2014large} (\S~\ref{sec:compari_sota}).} Notably, Zero123-NVS~\cite{liu2023zero123} only supports foreground objects, therefore, we mask out the foreground using their official code. Obviously, Zero123-NVS is sensitive to the estimated mask and shows significant inconsistency between different views with inaccurate geometry, while our method hierarchically includes the details in a coarse-to-fine manner with much better geometry and consistency. }
\label{fig_dtu_vis}
\end{figure*}

\vspace{0.1cm}
\noindent\textbf{Hyper Parameters.}
(i) \textit{OPP.} For the intermediate feature map $\textbf{I}_m$ in one-stage pipelines, we set $H_m=W_m=16$, $h_m=128$, \ie, $16\times16$ rays for grid sampling. The resolution of the random patch sampling is also set as $16\times16$. 
For fair comparisons, we set the same training hyperparameters for all the pipelines. Specifically, for the generative learning objectives, we set $\lambda_{GAN}=1e^{-3}$ and $\lambda_{PER}=1e^{-2}$. To rigorously validate the effectiveness of our proposed method, we strictly keep the learning strategy of PixelNeRF such as the learning rate ($1e^{-4}$), optimizer (Adam solver~\cite{Adam_kingma2014adam}), and sampling points per ray ($64$ coarse and $96$ fine).

(ii) \textit{Diff3DE.} The dense keyframe number $N_k$ is set as $40$, which uniformly dispersed around the dome. 
We resize the rendered images from the OPP module to $512\times512$ resolution as the Diff3DE inputs. The classifier-free guidance scale is set as $7.5$ and the denoising step number is $25$. 
For the ShapeNet Cars and Chairs categories, we use the simple text descriptions 'A car' and 'A chair', respectively; and for the DTU dataset, we use~\cite{li2023blip2}  to generate text descriptions.

\subsection{\textbf{Comparison with State-of-the-art}}
\label{sec:compari_sota}

\noindent\textbf{Out-distribution Baselines.} Since the main goal of final fine stage is achieving vivid plausible details with 3D consistency in an inference-time finetuning-free manner via out-distribution priors, we choose recent Zero-123-NVS~\cite{liu2023zero123} as our main competitor, which is a finetuned latent diffusion model with viewpoint condition and can also synthesize novel views without per-scene finetuning using out-distribution priors.  

\vspace{0.1cm}
\noindent\textbf{In-distribution Baselines.} The main in-distribution baseline of our method is PixelNeRF~\cite{PixelNeRF_yu2021pixelnerf}, which is taken as the OG-NeRF part of our GD$^2$-NeRF. Additionally, we report the quantitative results of methods including 3D GAN methods~\cite{PiGAN_chan2021pi,Pix2NeRF_cai2022pix2nerf,chan2022EG3D}, Geometry-free method~\cite{watson20223DiM},  Large-model-based Image-to-3D methods~\cite{Jain_2021_ICCV_dietnerf,Xu_2022_SinNeRF}, and previous OG-NeRF methods~\cite{SRN_sitzmann2019scene,CodeNeRF_jang2021codenerf,FE_NVS_guo2022fast}. Notably, VisionNeRF~\cite{lin2023visionnerf} and NeRFDiff~\cite{gu2023nerfdiff} are not listed here since they require much more computation than PixelNeRF~\cite{lin2023visionnerf,gu2023nerfdiff} (detailed in the appendix), and ~\cite{gu2023nerfdiff} requires tedious per-scene optimization using the co-trained diffusion model.

\noindent\textbf{Qualitative Analysis.}
We perform qualitative comparisons with previous methods on ShapNet Cars \& Chairs in Fig.~\ref{fig_srn_nmr_sota}, and DTU dataset in Fig.~\ref{fig_dtu_vis}. Obviously, Zero123-NVS shows significant inconsistency between different views together with inaccurate geometries. Also, it is sensitive to the foreground masking process, which is not suitable for relatively complex scenes as in the DTU dataset. In contrast, our method can gradually include the in- and out-distribution details while maintaining good 3D consistency and geometry on both synthetic and real-world complex datasets. 

\noindent\textbf{Quantitative Analysis.}
(i) \textit{Out-distribution 3D-consistency.}
We compare the 3D-consistency with our main out-distribution competitor, Zero123-NVS, quantitatively in Tab.~\ref{tab:3d_consistency} using $10$ randomly picked videos from each dataset, where our method outperforms Zero123-NVS in Pixel-MSE score by almost twice. This can be credited to the primary consistency provided by the coarse-stage outputs together with the proposed global consistency approximation in fine-stage Diff3DE. 
\\
\noindent (ii) \textit{In-distribution Image Quality.}
We report the in-distribution metric comparisons in Tab.~\ref{tab:category_specific}. It is obvious that our in-distribution method OPP shows a good balance between sharpness and fidelity in general, outperforming the baseline methods even though many of them require per-scene optimization/finetuning/auto-regression. It is worth mentioning that though 3DiM~\cite{watson20223DiM} achieves the best FID, as a geometry-free method, it mainly focuses on the image quality of every single view while ignoring the 3D consistency.  

\begin{table}[t]
\setlength{\abovecaptionskip}{0cm}
\footnotesize
    \caption{\textbf{Out-distribution comparisons between our fine-stage method Diff3DE and Zero123-NVS on ShapeNet and DTU (\S~\ref{sec:compari_sota}).} The 3D-consistency of our Diff3DE outperforms Zero123-NVS by nearly two times. }
    \label{tab:3d_consistency}
    \setlength\tabcolsep{7.5pt}
    


\begin{tabular}{l|ccc}
 \rowcolor[gray]{.9}
\hline
                & \multicolumn{3}{c}{$\downarrow${Pixel-MSE}}                            \\  \rowcolor[gray]{.9}
{Methods} & {ShapeNet Cars} & {ShapeNet Chairs} & {DTU}     \\

 \hline \hline
Zero123-NVS         & 939.67                 & 642.67                   & 4350.70          \\
\textbf{Diff3DE (Ours)}         & \cb{489.30}        & \cb{330.81}          & \cb{2352.69} \\ \hline
\end{tabular}

\end{table} 

\begin{table}[t]
    \setlength{\abovecaptionskip}{0cm}
    \footnotesize

     \caption{\textbf{In-distribution comparisons between our coarse-stage method OPP and previous methods on ShapeNet and DTU (\S~\ref{sec:compari_sota}).} We achieve improvements with balanced fidelity (PSNR, SSIM) and sharpness (LPIPS, FID). ``\dag'' means per-scene optimization/finetuning/auto-regression is required.  ``*'' indicates FID in $64 \times 64$ resolution.}
     \label{tab:category_specific}

\setlength\tabcolsep{5.2pt}

\begin{tabular}{lccccc}
\rowcolor[gray]{.9}
\hline
 & \multicolumn{2}{c}{Fidelity}       & \multicolumn{3}{c}{Sharpness}                       \\
 \rowcolor[gray]{.9}
Methods     & $\uparrow$ PSNR      & $\uparrow$ SSIM        & $\downarrow$ LPIPS      & $\downarrow$ FID  & $\downarrow$ FID$^*$

\\ \hline\hline
\multicolumn{6}{c}{\textit{ShapeNet  Chairs}} \\

$\pi$-GAN~\cite{PiGAN_chan2021pi}~\dag     & - & -               & -             & -                     &  15.47   \\                                
Pix2NeRF~\cite{Pix2NeRF_cai2022pix2nerf}  & {18.14}  & 0.84             & -   & -                                       &  \cs{14.31}   \\
        
SRN~\cite{SRN_sitzmann2019scene}~\dag         & 22.89       & 0.89                           & 0.104                      &  -       & -          \\
CodeNeRF~\cite{CodeNeRF_jang2021codenerf}~\dag                       & 22.39                           & 0.87                           & 0.166          &    -  & - \\
FE-NVS~\cite{FE_NVS_guo2022fast}                & 23.21                           & \cb{0.92}                           & \cs{0.077 }                     &    -   & -         \\ 
PixelNeRF~\cite{PixelNeRF_yu2021pixelnerf}     & \cs{23.72}                           & \cs{0.91}                           & 0.128               & 38.49    & -        \\ 
\textbf{OPP (ours)} & \cb{24.03}      & \cb{0.92}     & \cb{0.067} &   \cb{15.10}  & \cb{7.86} \\ 
                         
                \hline \hline

\multicolumn{6}{c}{\textit{ShapeNet  Cars}} \\

EG3D-PTI~\cite{chan2022EG3D}~\dag & 19.00	& 0.85	& 0.150	 & \cs{27.32}  & - \\

3DiM~\cite{watson20223DiM}~\dag   & 21.01 & 0.57	& -	& \cb{8.99}  & - \\


SRN~\cite{SRN_sitzmann2019scene}~\dag      & 22.25                           & 0.89                           & 0.129                      &    -      & -   \\

CodeNeRF~\cite{CodeNeRF_jang2021codenerf}~\dag                       & 22.73                           & 0.89                           & 0.128                      &       -       & -      \\
FE-NVS~\cite{FE_NVS_guo2022fast}                         & 22.83                           & \cb{0.91} & \cs{0.099 }                     &            -  & -     \\
 PixelNeRF~\cite{PixelNeRF_yu2021pixelnerf}                      & \cs{23.17} & \cs{0.90}                           & 0.146                      &   59.15         & -    \\ 
 \textbf{OPP (ours)} & \cb{23.24}      &  \cb{0.91}     & \cb{0.092} &    {33.53}  & -    

\\ \hline \hline

\multicolumn{6}{c}{\textit{DTU Dataset}} \\ 

    DietNeRF~\cite{Jain_2021_ICCV_dietnerf}~\dag	& 14.24 &	0.481	& 0.487	& 190.7 & - \\
    PixelNeRF~\cite{PixelNeRF_yu2021pixelnerf}   & {15.55}                         & {0.537}                    & 0.535             &  - & -  \\
        \textbf{OPP (ours)} & \cb{16.51}      &  \cb{0.659}     & \cb{0.399} & \cb{146.56}  & - \\ \hline

        \multicolumn{6}{c}{\textit{DTU Dataset (SinNeRF Split)}} \\

    SinNeRF~\cite{Xu_2022_SinNeRF}~\dag & 11.18	& 0.424 &	0.571	& 283.86 & - \\
    \textbf{OPP (ours)} & \cb{17.27}	& \cb{0.730}	& \cb{0.354}	& \cb{146.30} & - \\ \hline

\end{tabular}


\end{table}

\subsection{\textbf{Ablation Studies of Out-distribution Diff3DE}}
\label{sec:ablation_Diff3DE}

\noindent\textbf{Effectiveness of Coarse-to-Fine.}
\begin{figure*}[t]
\small
\centering
\includegraphics[width=0.85\linewidth]{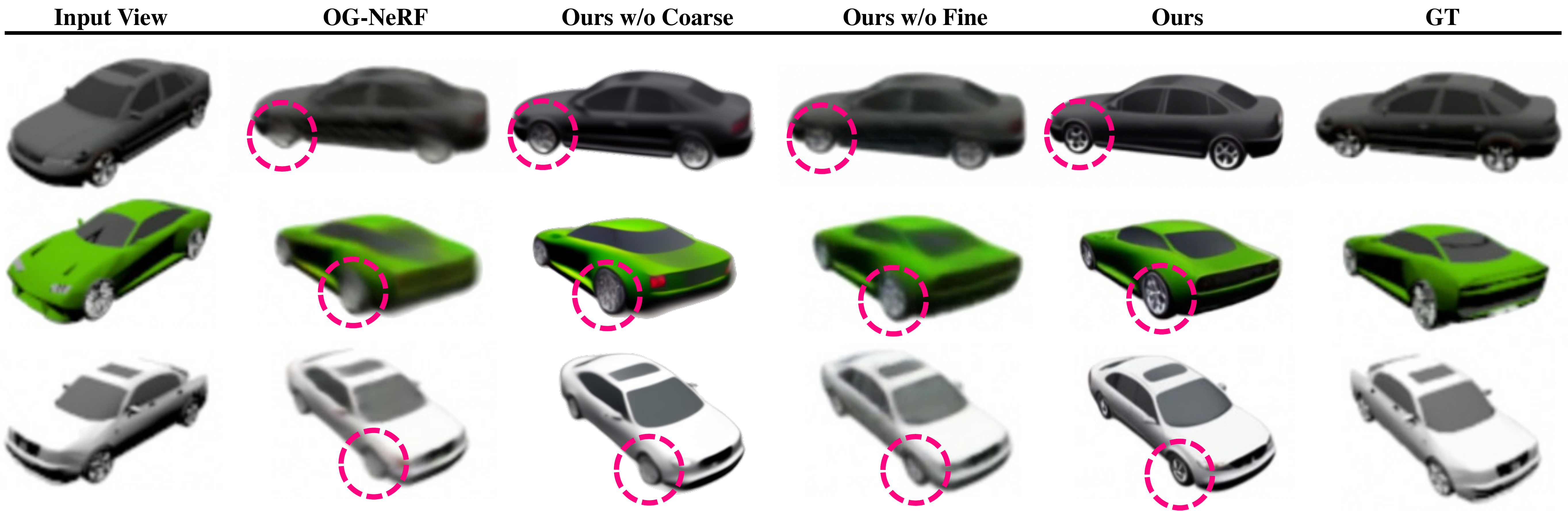}
\caption{\textbf{Effectiveness of coarse-to-fine (\S~\ref{sec:ablation_Diff3DE}).} ``w/o Coarse'' indicates directly adding Diff3DE on the original blurry OG-NeRF, and ``w/o Fine'' means only using OPP. Obviously, our proposed coarse-to-fine scheme gradually includes the details and gives the best results.}
\label{fig_ablation_c2f}
\end{figure*}
We verify the effectiveness of the coarse-to-fine strategy in Fig.~\ref{fig_ablation_c2f}. When directly adding Diff3DE on the blurry OG-NeRF outputs (``Ours w/o Coarse''), the results tend to lose rich details and even significantly wrong geometry, \eg, the window of the green car. With the proposed coarse-to-fine method, the wrong geometry can be generally corrected and the details are gradually included, formulating vivid outputs.

\noindent\textbf{Influence of Dense Keyframe Number $N_k$.}
\begin{figure}[t]
\small
\centering
\includegraphics[width=1.0\linewidth]{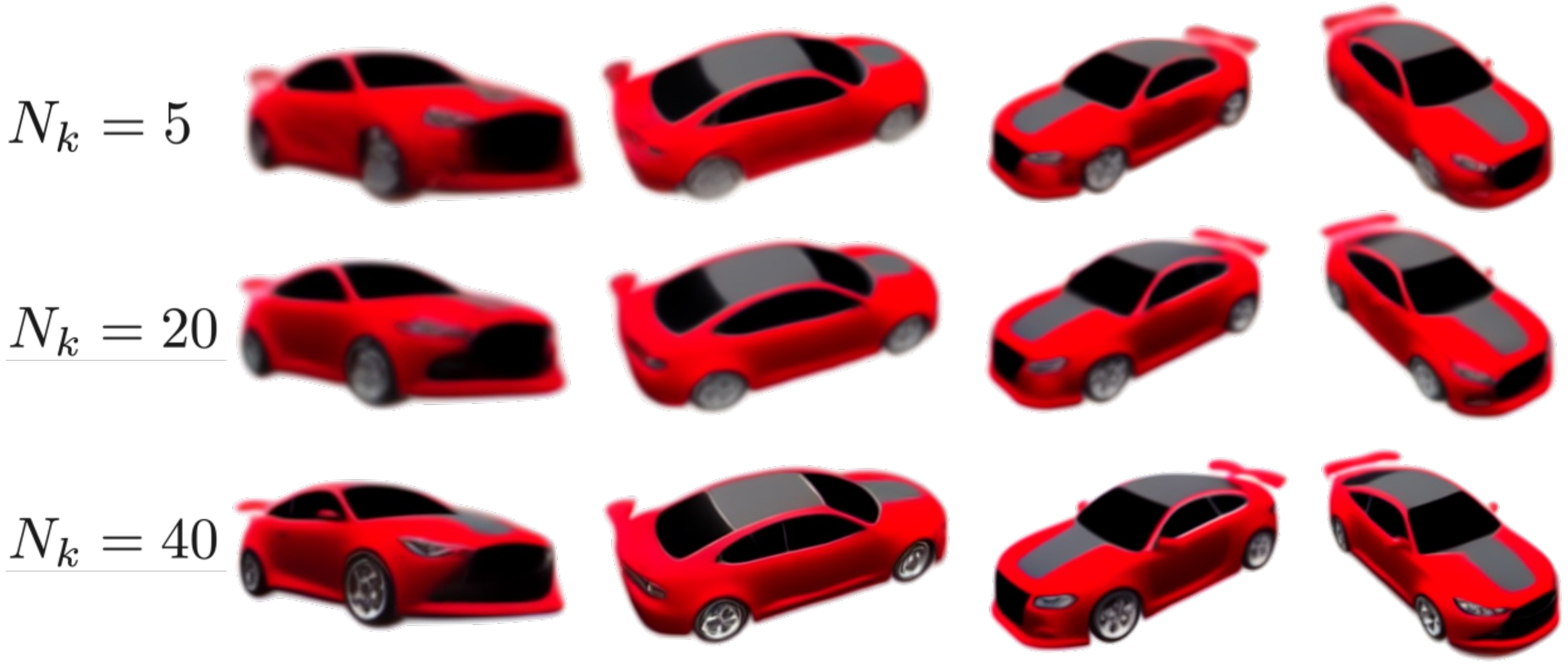}
\caption{\textbf{Influence of the dense key frame number $N_k$ (\S~\ref{sec:ablation_Diff3DE}).} Small $N_k$ leads to more dispersed keyframes with large content variance, therefore generating blurry outputs. Increasing $N_k$ relieves this issue via enabling more similar content in neighbor sets.}
\label{fig_ablation_Nk}
\end{figure}
The influence of dense keyframe number $N_k$ is illustrated in Fig.~\ref{fig_ablation_Nk}. With small $N_k$, the input of ISA module tends to contain contents with large variance, therefore leads to blurry outputs. Since our Diff3DE takes neighbor keyframe sets as ISA inputs, we are able to increase the dense keyframe number $N_k$ to relieve the content variance of ISA inputs. Obviously, the blurry issue is relieved as $N_k$ increases.

\subsection{\textbf{Ablation Studies of In-distribution OPP}}

\label{sec:ablation_OPP}
Similar to~\cite {PixelNeRF_yu2021pixelnerf}, we perform in-distribution ablation studies with 10\% random instances (fixed across all experiments) from the test set of the ShapeNet Cars category. We provide the efficiency analysis of OPP in the appendix.

\vspace{0.1cm}
\noindent\textbf{Analysis of Pipelines.}
We report the performance of different pipelines in Table \ref{tab:ablation_pipelines}. Compared with the independently trained OG-NeRF, \ie, PixelNeRF, the Two-stage Tandem Pipeline (TTP) improves the sharpness, \eg, the FID score is decreased by $21.63$, yet significantly suffers the loss of fidelity, \eg, $-1.30$ in PSNR score. With the end-to-end training in the One-stage Tandem Pipeline (OTP), the fidelity is greatly improved, but there still exist large margins compared with the independently trained OG-NeRF, \eg, $23.11$ \vs \  $23.61$ in PSNR score. In contrast, our final One-stage Parallel Pipeline (OPP) achieves both high fidelity and sharpness.

\begin{table}[t]
\setlength{\abovecaptionskip}{0cm}
\footnotesize
     \caption{\textbf{ Ablation of different pipelines (\S~\ref{sec:ablation_OPP}).} Compared with the independently trained OG-NeRF model and two basic tandem pipelines: Two-stage Tandem Pipeline (TTP) and One-stage Tandem Pipeline (OTP), our final One-stage Parallel Pipeline (OPP) achieves more balanced fidelity and sharpness.}
\label{tab:ablation_pipelines}

    \setlength\tabcolsep{11pt}
    \begin{tabular}{l|cccc}
    \rowcolor[gray]{.9}
    \hline
     & \multicolumn{2}{c}{Fidelity}       & \multicolumn{2}{c}{Sharpness}                       \\
 \rowcolor[gray]{.9}
    Methods           & $\uparrow$ PSNR                & $\uparrow$ SSIM               & $\downarrow$ LPIPS      & $\downarrow$ FID \\ \hline \hline

    OG-NeRF                   & \cs{23.61}                         & \cs{0.907}                           & 0.113                      &   69.85    \\
    TTP       &    22.31                        &     0.899                      &        \cs{0.089}        &  48.22   \\ 
    OTP         &  23.11                &    0.900                       &   \cb{0.086}        &  \cs{40.00}                 \\ 

    \textbf{OPP (ours)} & \cb{23.69}      & \cb{0.910}     & {0.091} & \cb{39.70} \\ \hline
    \end{tabular}

\end{table}
\begin{table}[t]
\setlength{\abovecaptionskip}{0cm}
\footnotesize
    \caption{\textbf{Effectiveness of DPS (\S~\ref{sec:ablation_OPP}).} Compared with the independently trained generative and OG-NeRF models, the jointly optimized ones from DPS can achieve better performance. }
    \label{tab:ablation_dps}
    \setlength\tabcolsep{8.6pt}
        \begin{tabular}{l|cccc}
        \rowcolor[gray]{.9}
        \hline
         & \multicolumn{2}{c}{Fidelity}       & \multicolumn{2}{c}{Sharpness}                       \\
 \rowcolor[gray]{.9}
        Methods           & $\uparrow$ PSNR                & $\uparrow$ SSIM               & $\downarrow$ LPIPS      & $\downarrow$ FID \\ \hline \hline

        Generative  &  23.61   & 0.907           & 0.113            & {69.85}   \\
        \textbf{Generative (DPS)}  &  \cb{ 23.64 }                     &     \cb{ 0.909}              &    \cb{0.108}         &   \cb{61.07}  \\ \hline  \hline
        OG-NeRF  &    23.11                &  0.900                  &   0.086          &  \cb{40.00}  \\
        \textbf{OG-NeRF (DPS)} &  \cb{23.16}                      &    \cb{0.902}                &  \cb{0.085}           &  41.35  \\ 
        \hline
        \end{tabular}

\end{table}


\begin{table}[t]
\setlength{\abovecaptionskip}{0cm}
\footnotesize
     \caption{\textbf{ Effectiveness of fusing via the output confidence map of CoRF (\S~\ref{sec:ablation_OPP}).} The fused one mines the benefits of both the OG-NeRF (PSNR, SSIM) and generative model (LPIPS, FID).  }
    \label{tab:ablation_rvba}
    \setlength\tabcolsep{8.5pt}
        \begin{tabular}{l|cccc}
        \rowcolor[gray]{.9}
        \hline
         & \multicolumn{2}{c}{Fidelity}       & \multicolumn{2}{c}{Sharpness}                       \\
 \rowcolor[gray]{.9}
        Methods           & $\uparrow$ PSNR                & $\uparrow$ SSIM               & $\downarrow$ LPIPS      & $\downarrow$ FID \\ \hline \hline

        OG-NeRF (DPS)   & {\cs{23.64}}                         & {\cs{0.909}}                    & 0.108             &  61.07   \\
        Generative (DPS)   &  23.16                        & 0.902                    & \cb{0.085}             &  \cs{41.35}   \\

        \textbf{CoRF + DPF} & \cb{23.69}      & \cb{0.910}     & {\cs{0.091}} & \cb{39.70} \\ 
        \hline
        \end{tabular}

\end{table}


\begin{figure}[h]
\setlength{\abovecaptionskip}{0cm}
\small
\centering
\includegraphics[width=1.0\linewidth]{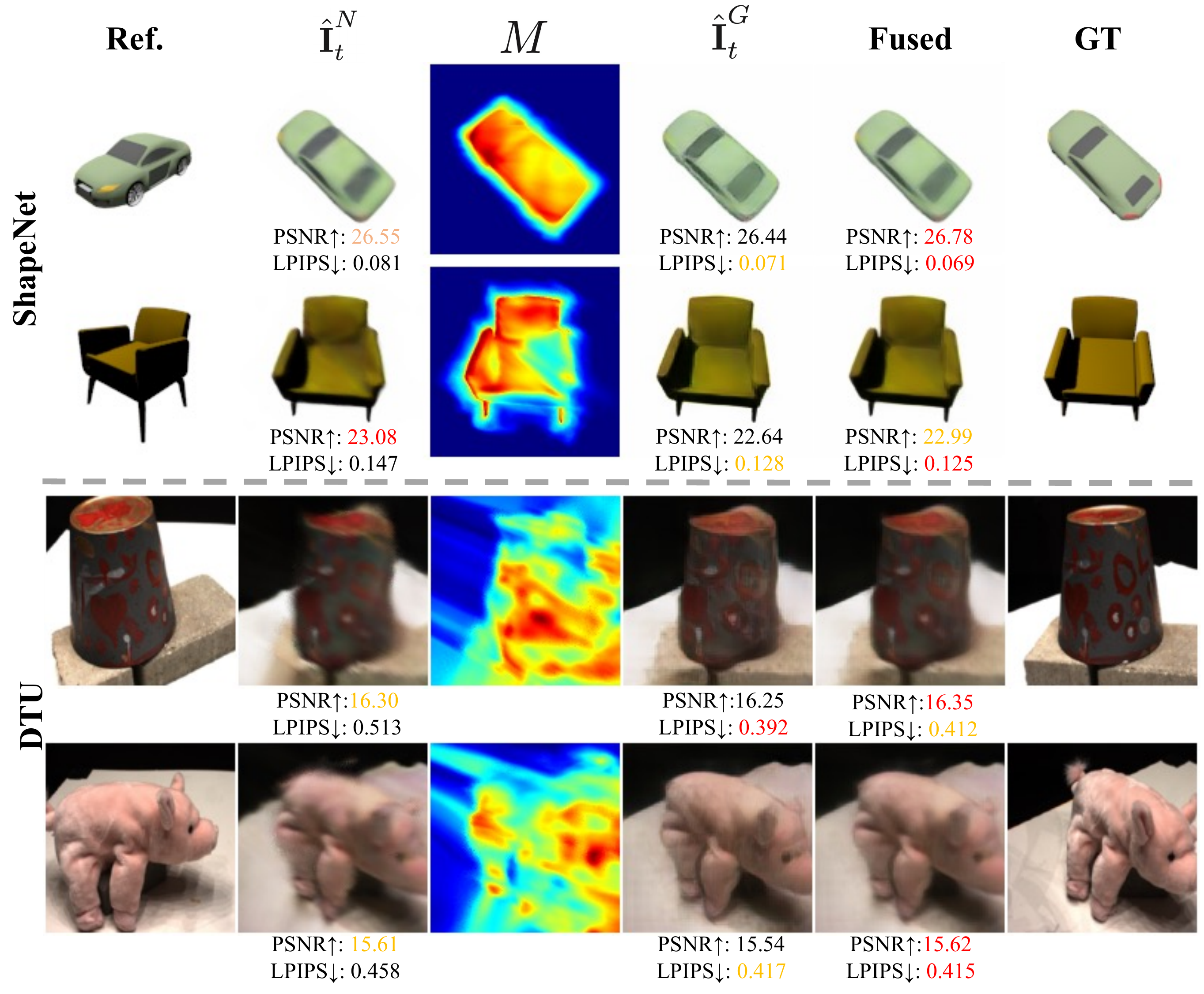}
\caption{\textbf{Visualization of the CoRF-predicted confidence map  and the DPF results on ShapeNet and DTU (\S~\ref{sec:ablation_OPP}).} CoRF successfully learns to adaptively give the blurry part with low confidence, and the fused ones achieve good balance between two paradigms. Best viewed with zoom-in for details.}
\label{fig_corf_vis}
\end{figure}

    

\vspace{0.1cm}
\noindent\textbf{Effectiveness of DPS.}
We study the influence of jointly optimizing OG-NeRF and generative models with DPS in Table \ref{tab:ablation_dps}.  Compared with the independently trained OG-NeRF (\ie, PixelNeRF) and generative models (\ie, one-stage tandem pipeline), we find the DPS outputs give better performance, which proves that DPS is a lightweight yet effective structure that can facilitate the parallel optimization of OG-NeRF and generative models.

\vspace{0.1cm}
\noindent\textbf{Effectiveness of CoRF \& DPF.} Table~\ref{tab:ablation_rvba} proves the effectiveness of fusing with the output confidence map of CoRF, where the fused one achieves both high fidelity and sharpness. We also illustrate the learned confidence map in Fig.~\ref{fig_corf_vis}. Obviously, CoRF successfully learns to give the blurry part with lower confidence,  and the fused ones can achieve a good balance between $\hat{\textbf{I}}_t^{N}$ and  $\hat{\textbf{I}}_t^{G}$.

\section{Conclusion}
In this paper, targeting the blurry issue of existing OG-NeRF methods, we propose the GD$^2$-NeRF, a generative detail compensation framework via GAN and pre-trained diffusion models for O-NVS task. It achieves vivid plausible outputs in an inference-time finetuning-free manner with decent 3D-consistency. We hope that our efforts will motivate more researchers in the future. 

\noindent\textbf{Limitations.}  
First, same as most diffusion-based methods~\cite{gu2023nerfdiff,yang2023rerender,geyer2023tokenflow}, the denoising process is inefficient. Second, as mentioned in~\cite{geyer2023tokenflow}, the decoder of the latent diffusion model introduces several high frequency flickering, which can be possibly addressed via an improved decoder or existing post-process deflickering method~\cite{geyer2023tokenflow}. Third, our fine-stage method Diff3DE complements details mainly on the basis of the input contents, and cannot correct large geometry artifacts.

\bibliographystyle{IEEEtran}
\bibliography{./ref}

\begin{thebibliography}{10}
\providecommand{\url}[1]{#1}
\csname url@samestyle\endcsname
\providecommand{\newblock}{\relax}
\providecommand{\bibinfo}[2]{#2}
\providecommand{\BIBentrySTDinterwordspacing}{\spaceskip=0pt\relax}
\providecommand{\BIBentryALTinterwordstretchfactor}{4}
\providecommand{\BIBentryALTinterwordspacing}{\spaceskip=\fontdimen2\font plus
\BIBentryALTinterwordstretchfactor\fontdimen3\font minus \fontdimen4\font\relax}
\providecommand{\BIBforeignlanguage}[2]{{%
\expandafter\ifx\csname l@#1\endcsname\relax
\typeout{** WARNING: IEEEtran.bst: No hyphenation pattern has been}%
\typeout{** loaded for the language `#1'. Using the pattern for}%
\typeout{** the default language instead.}%
\else
\language=\csname l@#1\endcsname
\fi
#2}}
\providecommand{\BIBdecl}{\relax}
\BIBdecl

\bibitem{PixelNeRF_yu2021pixelnerf}
A.~Yu, V.~Ye, M.~Tancik, and A.~Kanazawa, ``pixelnerf: Neural radiance fields from one or few images,'' in \emph{Proceedings of the IEEE/CVF Conference on Computer Vision and Pattern Recognition}, 2021, pp. 4578--4587.

\bibitem{rombach2022LatentDiff}
R.~Rombach, A.~Blattmann, D.~Lorenz, P.~Esser, and B.~Ommer, ``High-resolution image synthesis with latent diffusion models,'' in \emph{Proceedings of the IEEE/CVF conference on computer vision and pattern recognition}, 2022, pp. 10\,684--10\,695.

\bibitem{zhang2023addingControlNet}
L.~Zhang, A.~Rao, and M.~Agrawala, ``Adding conditional control to text-to-image diffusion models,'' 2023.

\bibitem{lin2023visionnerf}
K.-E. Lin, L.~Yen-Chen, W.-S. Lai, T.-Y. Lin, Y.-C. Shih, and R.~Ramamoorthi, ``Vision transformer for nerf-based view synthesis from a single input image,'' in \emph{WACV}, 2023.

\bibitem{FE_NVS_guo2022fast}
P.~Guo, M.~A. Bautista, A.~Colburn, L.~Yang, D.~Ulbricht, J.~M. Susskind, and Q.~Shan, ``Fast and explicit neural view synthesis,'' in \emph{Proceedings of the IEEE/CVF Winter Conference on Applications of Computer Vision}, 2022, pp. 3791--3800.

\bibitem{liu2023zero123}
R.~Liu, R.~Wu, B.~Van~Hoorick, P.~Tokmakov, S.~Zakharov, and C.~Vondrick, ``Zero-1-to-3: Zero-shot one image to 3d object,'' in \emph{Proceedings of the IEEE/CVF International Conference on Computer Vision}, 2023, pp. 9298--9309.

\bibitem{tang2023makeit3d}
J.~Tang, T.~Wang, B.~Zhang, T.~Zhang, R.~Yi, L.~Ma, and D.~Chen, ``Make-it-3d: High-fidelity 3d creation from a single image with diffusion prior,'' \emph{arXiv preprint arXiv:2303.14184}, 2023.

\bibitem{yang2023rerender}
S.~Yang, Y.~Zhou, Z.~Liu, and C.~C. Loy, ``Rerender a video: Zero-shot text-guided video-to-video translation,'' \emph{arXiv preprint arXiv:2306.07954}, 2023.

\bibitem{geyer2023tokenflow}
M.~Geyer, O.~Bar-Tal, S.~Bagon, and T.~Dekel, ``Tokenflow: Consistent diffusion features for consistent video editing,'' \emph{arXiv preprint arXiv:2307.10373}, 2023.

\bibitem{gu2023nerfdiff}
J.~Gu, A.~Trevithick, K.-E. Lin, J.~Susskind, C.~Theobalt, L.~Liu, and R.~Ramamoorthi, ``Nerfdiff: Single-image view synthesis with nerf-guided distillation from 3d-aware diffusion,'' in \emph{International Conference on Machine Learning}, 2023.

\bibitem{sitzmann2021LFN}
V.~Sitzmann, S.~Rezchikov, B.~Freeman, J.~Tenenbaum, and F.~Durand, ``Light field networks: Neural scene representations with single-evaluation rendering,'' \emph{Advances in Neural Information Processing Systems}, vol.~34, pp. 19\,313--19\,325, 2021.

\bibitem{srt22}
\BIBentryALTinterwordspacing
M.~S.~M. Sajjadi, H.~Meyer, E.~Pot, U.~Bergmann, K.~Greff, N.~Radwan, S.~Vora, M.~Lucic, D.~Duckworth, A.~Dosovitskiy, J.~Uszkoreit, T.~Funkhouser, and A.~Tagliasacchi, ``{Scene Representation Transformer: Geometry-Free Novel View Synthesis Through Set-Latent Scene Representations},'' \emph{{CVPR}}, 2022. [Online]. Available: \url{https://srt-paper.github.io/}
\BIBentrySTDinterwordspacing

\bibitem{watson20223DiM}
\BIBentryALTinterwordspacing
D.~Watson, W.~Chan, R.~M. Brualla, J.~Ho, A.~Tagliasacchi, and M.~Norouzi, ``Novel view synthesis with diffusion models,'' in \emph{The Eleventh International Conference on Learning Representations}, 2023. [Online]. Available: \url{https://openreview.net/forum?id=HtoA0oT30jC}
\BIBentrySTDinterwordspacing

\bibitem{PiGAN_chan2021pi}
E.~R. Chan, M.~Monteiro, P.~Kellnhofer, J.~Wu, and G.~Wetzstein, ``pi-gan: Periodic implicit generative adversarial networks for 3d-aware image synthesis,'' in \emph{Proceedings of the IEEE/CVF conference on computer vision and pattern recognition}, 2021, pp. 5799--5809.

\bibitem{Pix2NeRF_cai2022pix2nerf}
S.~Cai, A.~Obukhov, D.~Dai, and L.~Van~Gool, ``Pix2nerf: Unsupervised conditional p-gan for single image to neural radiance fields translation,'' in \emph{Proceedings of the IEEE/CVF Conference on Computer Vision and Pattern Recognition}, 2022, pp. 3981--3990.

\bibitem{chan2022EG3D}
E.~R. Chan, C.~Z. Lin, M.~A. Chan, K.~Nagano, B.~Pan, S.~De~Mello, O.~Gallo, L.~J. Guibas, J.~Tremblay, S.~Khamis \emph{et~al.}, ``Efficient geometry-aware 3d generative adversarial networks,'' in \emph{Proceedings of the IEEE/CVF Conference on Computer Vision and Pattern Recognition}, 2022, pp. 16\,123--16\,133.

\bibitem{Xu_2022_SinNeRF}
D.~Xu, Y.~Jiang, P.~Wang, Z.~Fan, H.~Shi, and Z.~Wang, ``Sinnerf: Training neural radiance fields on complex scenes from a single image,'' 2022.

\bibitem{Jain_2021_ICCV_dietnerf}
A.~Jain, M.~Tancik, and P.~Abbeel, ``Putting nerf on a diet: Semantically consistent few-shot view synthesis,'' in \emph{Proceedings of the IEEE/CVF International Conference on Computer Vision (ICCV)}, October 2021, pp. 5885--5894.

\bibitem{NeRF_mildenhall2020nerf}
B.~Mildenhall, P.~P. Srinivasan, M.~Tancik, J.~T. Barron, R.~Ramamoorthi, and R.~Ng, ``Nerf: Representing scenes as neural radiance fields for view synthesis,'' in \emph{European conference on computer vision}.\hskip 1em plus 0.5em minus 0.4em\relax Springer, 2020, pp. 405--421.

\bibitem{CodeNeRF_jang2021codenerf}
W.~Jang and L.~Agapito, ``Codenerf: Disentangled neural radiance fields for object categories,'' in \emph{Proceedings of the IEEE/CVF International Conference on Computer Vision}, 2021, pp. 12\,949--12\,958.

\bibitem{ShaRF_rematas2021sharf}
K.~Rematas, R.~Martin-Brualla, and V.~Ferrari, ``Sharf: Shape-conditioned radiance fields from a single view,'' \emph{arXiv preprint arXiv:2102.08860}, 2021.

\bibitem{GAN_resolution_brock2018large}
A.~Brock, J.~Donahue, and K.~Simonyan, ``Large scale gan training for high fidelity natural image synthesis,'' \emph{arXiv preprint arXiv:1809.11096}, 2018.

\bibitem{GAN_resolution_choi2018stargan}
Y.~Choi, M.~Choi, M.~Kim, J.-W. Ha, S.~Kim, and J.~Choo, ``Stargan: Unified generative adversarial networks for multi-domain image-to-image translation,'' in \emph{Proceedings of the IEEE conference on computer vision and pattern recognition}, 2018, pp. 8789--8797.

\bibitem{GAN_resolution_karras2019style}
T.~Karras, S.~Laine, and T.~Aila, ``A style-based generator architecture for generative adversarial networks,'' in \emph{Proceedings of the IEEE/CVF conference on computer vision and pattern recognition}, 2019, pp. 4401--4410.

\bibitem{GRAF_schwarz2020graf}
K.~Schwarz, Y.~Liao, M.~Niemeyer, and A.~Geiger, ``Graf: Generative radiance fields for 3d-aware image synthesis,'' \emph{Advances in Neural Information Processing Systems}, vol.~33, pp. 20\,154--20\,166, 2020.

\bibitem{GIRAFFE_niemeyer2021giraffe}
M.~Niemeyer and A.~Geiger, ``Giraffe: Representing scenes as compositional generative neural feature fields,'' in \emph{Proceedings of the IEEE/CVF Conference on Computer Vision and Pattern Recognition}, 2021, pp. 11\,453--11\,464.

\bibitem{caron2021emerging}
M.~Caron, H.~Touvron, I.~Misra, H.~J{\'e}gou, J.~Mairal, P.~Bojanowski, and A.~Joulin, ``Emerging properties in self-supervised vision transformers,'' in \emph{Proceedings of the IEEE/CVF international conference on computer vision}, 2021, pp. 9650--9660.

\bibitem{radford2021CLIP}
A.~Radford, J.~W. Kim, C.~Hallacy, A.~Ramesh, G.~Goh, S.~Agarwal, G.~Sastry, A.~Askell, P.~Mishkin, J.~Clark \emph{et~al.}, ``Learning transferable visual models from natural language supervision,'' in \emph{International conference on machine learning}.\hskip 1em plus 0.5em minus 0.4em\relax PMLR, 2021, pp. 8748--8763.

\bibitem{deng2023nerdi}
C.~Deng, C.~Jiang, C.~R. Qi, X.~Yan, Y.~Zhou, L.~Guibas, D.~Anguelov \emph{et~al.}, ``Nerdi: Single-view nerf synthesis with language-guided diffusion as general image priors,'' in \emph{Proceedings of the IEEE/CVF Conference on Computer Vision and Pattern Recognition}, 2023, pp. 20\,637--20\,647.

\bibitem{xu2022neurallift}
D.~Xu, Y.~Jiang, P.~Wang, Z.~Fan, Y.~Wang, and Z.~Wang, ``Neurallift-360: Lifting an in-the-wild 2d photo to a 3d object with 360 views,'' \emph{arXiv e-prints}, pp. arXiv--2211, 2022.

\bibitem{deitke2023objaverse}
M.~Deitke, D.~Schwenk, J.~Salvador, L.~Weihs, O.~Michel, E.~VanderBilt, L.~Schmidt, K.~Ehsani, A.~Kembhavi, and A.~Farhadi, ``Objaverse: A universe of annotated 3d objects,'' in \emph{Proceedings of the IEEE/CVF Conference on Computer Vision and Pattern Recognition}, 2023, pp. 13\,142--13\,153.

\bibitem{wang2023SJC}
H.~Wang, X.~Du, J.~Li, R.~A. Yeh, and G.~Shakhnarovich, ``Score jacobian chaining: Lifting pretrained 2d diffusion models for 3d generation,'' in \emph{Proceedings of the IEEE/CVF Conference on Computer Vision and Pattern Recognition}, 2023, pp. 12\,619--12\,629.

\bibitem{wu2023tune-a-video}
J.~Z. Wu, Y.~Ge, X.~Wang, S.~W. Lei, Y.~Gu, Y.~Shi, W.~Hsu, Y.~Shan, X.~Qie, and M.~Z. Shou, ``Tune-a-video: One-shot tuning of image diffusion models for text-to-video generation,'' in \emph{Proceedings of the IEEE/CVF International Conference on Computer Vision}, 2023, pp. 7623--7633.

\bibitem{text2video-zero}
L.~Khachatryan, A.~Movsisyan, V.~Tadevosyan, R.~Henschel, Z.~Wang, S.~Navasardyan, and H.~Shi, ``Text2video-zero: Text-to-image diffusion models are zero-shot video generators,'' \emph{arXiv preprint arXiv:2303.13439}, 2023.

\bibitem{ceylan2023pix2video}
D.~Ceylan, C.-H.~P. Huang, and N.~J. Mitra, ``Pix2video: Video editing using image diffusion,'' \emph{arXiv preprint arXiv:2303.12688}, 2023.

\bibitem{qi2023fatezero}
C.~Qi, X.~Cun, Y.~Zhang, C.~Lei, X.~Wang, Y.~Shan, and Q.~Chen, ``Fatezero: Fusing attentions for zero-shot text-based video editing,'' \emph{arXiv:2303.09535}, 2023.

\bibitem{max1995optical}
N.~Max, ``Optical models for direct volume rendering,'' \emph{IEEE Transactions on Visualization and Computer Graphics}, vol.~1, no.~2, pp. 99--108, 1995.

\bibitem{GAN_goodfellow2014generative}
I.~Goodfellow, J.~Pouget-Abadie, M.~Mirza, B.~Xu, D.~Warde-Farley, S.~Ozair, A.~Courville, and Y.~Bengio, ``Generative adversarial nets,'' \emph{Advances in neural information processing systems}, vol.~27, 2014.

\bibitem{R1reg_mescheder2018training}
L.~Mescheder, A.~Geiger, and S.~Nowozin, ``Which training methods for gans do actually converge?'' in \emph{International conference on machine learning}.\hskip 1em plus 0.5em minus 0.4em\relax PMLR, 2018, pp. 3481--3490.

\bibitem{Perceptual_wu2020unsupervised}
S.~Wu, C.~Rupprecht, and A.~Vedaldi, ``Unsupervised learning of probably symmetric deformable 3d objects from images in the wild,'' in \emph{Proceedings of the IEEE/CVF Conference on Computer Vision and Pattern Recognition}, 2020, pp. 1--10.

\bibitem{croitoru2023diffusion0}
F.-A. Croitoru, V.~Hondru, R.~T. Ionescu, and M.~Shah, ``Diffusion models in vision: A survey,'' \emph{IEEE Transactions on Pattern Analysis and Machine Intelligence}, 2023.

\bibitem{dhariwal2021diffusion1}
P.~Dhariwal and A.~Nichol, ``Diffusion models beat gans on image synthesis,'' \emph{Advances in neural information processing systems}, vol.~34, pp. 8780--8794, 2021.

\bibitem{song2020DDIM}
J.~Song, C.~Meng, and S.~Ermon, ``Denoising diffusion implicit models,'' \emph{arXiv preprint arXiv:2010.02502}, 2020.

\bibitem{Tumanyan_2023_CVPR_PnP}
N.~Tumanyan, M.~Geyer, S.~Bagon, and T.~Dekel, ``Plug-and-play diffusion features for text-driven image-to-image translation,'' in \emph{Proceedings of the IEEE/CVF Conference on Computer Vision and Pattern Recognition (CVPR)}, June 2023, pp. 1921--1930.

\bibitem{ShapeNet_chang2015shapenet}
A.~X. Chang, T.~Funkhouser, L.~Guibas, P.~Hanrahan, Q.~Huang, Z.~Li, S.~Savarese, M.~Savva, S.~Song, H.~Su \emph{et~al.}, ``Shapenet: An information-rich 3d model repository,'' \emph{arXiv preprint arXiv:1512.03012}, 2015.

\bibitem{SRN_sitzmann2019scene}
V.~Sitzmann, M.~Zollh{\"o}fer, and G.~Wetzstein, ``Scene representation networks: Continuous 3d-structure-aware neural scene representations,'' \emph{Advances in Neural Information Processing Systems}, vol.~32, 2019.

\bibitem{teed2020raft}
Z.~Teed and J.~Deng, ``Raft: Recurrent all-pairs field transforms for optical flow,'' in \emph{Computer Vision--ECCV 2020: 16th European Conference, Glasgow, UK, August 23--28, 2020, Proceedings, Part II 16}.\hskip 1em plus 0.5em minus 0.4em\relax Springer, 2020, pp. 402--419.

\bibitem{PSNR_wang2004image}
Z.~Wang, A.~C. Bovik, H.~R. Sheikh, and E.~P. Simoncelli, ``Image quality assessment: from error visibility to structural similarity,'' \emph{IEEE transactions on image processing}, vol.~13, no.~4, pp. 600--612, 2004.

\bibitem{LPIPS_zhang2018unreasonable}
R.~Zhang, P.~Isola, A.~A. Efros, E.~Shechtman, and O.~Wang, ``The unreasonable effectiveness of deep features as a perceptual metric,'' in \emph{Proceedings of the IEEE conference on computer vision and pattern recognition}, 2018, pp. 586--595.

\bibitem{FID_heusel2017gans}
M.~Heusel, H.~Ramsauer, T.~Unterthiner, B.~Nessler, and S.~Hochreiter, ``Gans trained by a two time-scale update rule converge to a local nash equilibrium,'' \emph{Advances in neural information processing systems}, vol.~30, 2017.

\bibitem{Unet_ronneberger2015u}
O.~Ronneberger, P.~Fischer, and T.~Brox, ``U-net: Convolutional networks for biomedical image segmentation,'' in \emph{International Conference on Medical image computing and computer-assisted intervention}.\hskip 1em plus 0.5em minus 0.4em\relax Springer, 2015, pp. 234--241.

\bibitem{DTU_jensen2014large}
R.~Jensen, A.~Dahl, G.~Vogiatzis, E.~Tola, and H.~Aan{\ae}s, ``Large scale multi-view stereopsis evaluation,'' in \emph{Proceedings of the IEEE conference on computer vision and pattern recognition}, 2014, pp. 406--413.

\bibitem{Adam_kingma2014adam}
D.~P. Kingma and J.~Ba, ``Adam: A method for stochastic optimization,'' \emph{arXiv preprint arXiv:1412.6980}, 2014.

\bibitem{li2023blip2}
J.~Li, D.~Li, S.~Savarese, and S.~Hoi, ``Blip-2: Bootstrapping language-image pre-training with frozen image encoders and large language models,'' \emph{arXiv preprint arXiv:2301.12597}, 2023.

\end{thebibliography}

\clearpage
\appendix

\subsection{Influence of $\lambda_{GAN}$ and $\lambda_{PER}$ in OPP}
\begin{table}[t]
\setlength{\abovecaptionskip}{0cm}
        \caption{\textbf{Influence of $\lambda_{GAN}$ and $\lambda_{PER}$ from generative learning objectives (Eq. \ref{eq:generative_objectives}).}}
    \label{tab:ablation_generative_objectives}
    
    \resizebox{\linewidth}{!}{
        \begin{tabular}{l|c|cccl}
        \rowcolor[gray]{.9}    
        \hline 
        Objectives   & Weight & $\uparrow$ PSNR & $\uparrow$ SSIM & $\downarrow$ LPIPS & $\downarrow$ FID \\ \hline \hline
        \multirow{3}{*}{$\lambda_{GAN}$} & $1e^{-4}$   &  23.49               &   0.906              &       0.097            &   42.43               \\
                                        & $1e^{-3}$  & \textbf{23.69}      & \textbf{0.910}     & \textbf{0.091} & \textbf{39.70}                  \\
                                        & $1e^{-2}$   & 23.62                 &    0.909             &     0.095               &  40.14                \\ \hline \hline
        \multirow{3}{*}{$\lambda_{PER}$} & $1e^{-3}$   &     23.64             &  0.908               &    0.105                &  51.21                \\
                                        & $1e^{-2}$  & \textbf{23.69}      & \textbf{0.910}     & \textbf{0.091} & \textbf{39.70}                  \\
                                       & $1e^{-1}$   &   23.28              &  0.902               &          0.100          &     46.70             \\ \hline
        \end{tabular}
    }
\end{table}
We study the influence of $\lambda_{GAN}$ and $\lambda_{PER}$ in Tab. \ref{tab:ablation_generative_objectives}. In line with Sec. \ref{sec:ablation_OPP}, we report the performance of the $10\%$ instances from the test set of the ShapeNet Cars category. We first fix $\lambda_{PER}$ as $1e^{-2}$ and then vary $\lambda_{GAN}$ as $\{1e^{-4}, 1e^{-3}, 1e^{-2}\}$. Obviously, $\lambda_{GAN} = 1e^{-3}$ gives the best performance. Then, we fix $\lambda_{GAN}$ as $1e^{-3}$, and vary $\lambda_{PER}$ as $\{1e^{-3}, 1e^{-2}, 1e^{-1}\}$, and find that $\lambda_{PER} = 1e^{-2}$ performs better.  We notice that the weight used in this work is much smaller than the previous GAN methods, \eg, $\lambda_{GAN}=\lambda_{PER}=5$ in \cite{Pix2NeRF_cai2022pix2nerf}. We infer that this is due to the functional difference of the GAN model between us. Specifically, in our work, the intermediate feature map $\textbf{I}_m$ already contains the coarse information provided by the OG-NeRF model, and the GAN model merely needs to refine the coarse input with learned prior. Therefore, too large weight will lead to over-imagination. While in \cite{Pix2NeRF_cai2022pix2nerf}, the captured prior by the GAN model is the main component for final outputs, therefore they require a larger weight.

\subsection{Efficiency Analysis of OPP}

\begin{table}[t]
\setlength{\abovecaptionskip}{0cm}
    \footnotesize
     \caption{\textbf{Efficiency comparisons with PixelNeRF.} The parameter number of OPP is only slightly higher than PixelNeRF, and the additional rendering of $\hat{\textbf{I}}^{G}$ and $M$ is significantly faster than $\hat{\textbf{I}}^{N}$. We test FPS on $1$ V100 with a batch size of $1$. }
    \label{tab:ablation_efficiency}
    \centering
    \small
     \setlength\tabcolsep{6.5pt}
        \begin{tabular}{l|cccc}
        \rowcolor[gray]{.9}
        \hline
        Methods           &  Param.   & FPS ($\hat{\textbf{I}}^{N}$) & FPS ($\hat{\textbf{I}}^{G}$)   & FPS ($M$)   \\ \hline \hline
        PixelNeRF~\cite{PixelNeRF_yu2021pixelnerf}   & 15.05M  &  0.54 &   -     &   -             \\
        \textbf{OPP (ours)} & 15.46M & 0.49  &  21.50 &  11.90 \\ 
        \hline
        \end{tabular}
\end{table}

\label{Sec:efficiency}
We report the efficiency comparisons with PixelNeRF in Table \ref{tab:ablation_efficiency}. With only an additional lightweight up-sampling module and several modifications on fully connected layers, the parameter number of our OPP is only $0.41M$ higher than PixelNeRF. During the inference stage, we need to forward two times, \ie, one full-sized ($H \times W$) rendering for getting $\hat{\textbf{I}}^{N}$ (same as PixelNeRF) and confidence map $M$, and one additional $H_m \times W_m$ rendering for $\hat{\textbf{I}}^{G}$. However, the CoRF MLP and up-sampling module are rather lightweight ($0.09M$, $0.11M$), and  $\hat{\textbf{I}}^{G}$ requires substantially fewer query points than $\hat{\textbf{I}}^N$. Therefore, rendering $\hat{\textbf{I}}^{G}$ and $M$ are over $40$ and $24$ times faster than $\hat{\textbf{I}}^{N}$ ($21.50$ \textit{vs.} $0.49$ FPS and $11.90$ \textit{vs.} $0.49$ FPS). In a nutshell, the inference speed is hardly affected.

\subsection{Comparisons of Computation Cost between OPP and Recent Works}

\begin{table}[t]
\setlength{\abovecaptionskip}{0cm}
    \footnotesize
     \caption{\textbf{Comparisons of computation cost with recent works~\cite{lin2023visionnerf,gu2023nerfdiff}.} Notably, ~\cite{gu2023nerfdiff} requires additional per-scene finetuning with unknown cost. }
    \label{tab:appendix_computationwithrecent}
    \centering
    \small
     \setlength\tabcolsep{8pt}
        \begin{tabular}{l|cc}
        \rowcolor[gray]{.9}
        \hline
        Methods    &  Param.   & Traning Time  \\ \hline \hline
        VisionNeRF~\cite{lin2023visionnerf}   & 122M  &  16*A100 5Days          \\
        NeRFDiff~\cite{gu2023nerfdiff}  & 400M / 1B &  8*A100 4Days + finetune    \\
        
        \textbf{OPP (ours)} & 15M & 8*V100  3Days \\ 
        \hline
        \end{tabular}
\end{table}
Recent works~\cite{lin2023visionnerf,gu2023nerfdiff} are not listed as competitors in Tab.~\ref{tab:category_specific} since they employ much heavier architecture with more computation cost, as illustrated in Tab.~\ref{tab:appendix_computationwithrecent}. Note that, except for the larger parameter number and training cost, NeRFDiff~\cite{gu2023nerfdiff} requires additional per-scene finetuning. In contrast, built on top of PixelNeRF, our OPP is quite lightweight, and requires no per-scene finetuning.






\vfill

\end{document}